\documentclass{article}
\usepackage{maitrix}

\usepackage{microtype}
\usepackage{graphicx}
\usepackage{booktabs} 

\usepackage{fancyhdr}
\usepackage{calc}
\usepackage{tcolorbox}
\usepackage{xspace}
\usepackage{titletoc}
\usepackage{wrapfig}
\usepackage{enumitem}

\usepackage{fontawesome5}  
\usepackage{listings}      
\usepackage[misc]{ifsym}
\usepackage{wrapfig}    
\usepackage[T1]{fontenc}
\usepackage{relsize}
\usepackage{adjustbox}
\usepackage{lipsum}
\usepackage{newtxtext}
\usepackage{algorithm}
\usepackage[noend]{algpseudocode}
\usepackage{ulem}
\usepackage{algorithm}
\usepackage{algpseudocode}  

\tcbuselibrary{skins, breakable}

\newlength\myheight
\newlength\mydepth
\settototalheight\myheight{Xygp}
\newcommand*\inlinegraphics[1]{%
  \settototalheight\myheight{Xygp}%
  \settodepth\mydepth{Xygp}%
  \raisebox{-1.6\mydepth}{\includegraphics[height=1.5\myheight]{#1}}%
}
\newcommand{\home}{\inlinegraphics{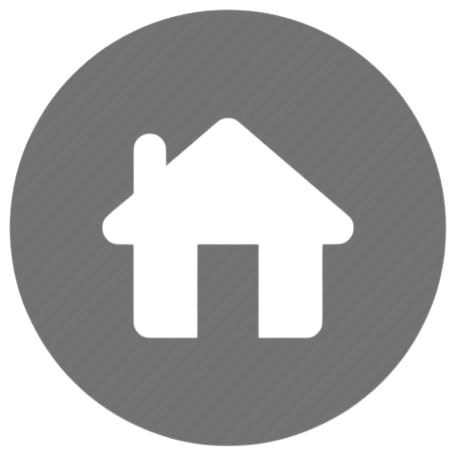}\xspace}

\definecolor{absbg}{RGB}{240,240,240}

\renewenvironment{abstract}{
  \begin{tcolorbox}[
    breakable,
    colback=absbg,
    colframe=white,
    boxrule=0pt,
    arc=12pt,
    left=8pt,
    right=8pt,
    top=8pt,
    bottom=8pt,
    width=\textwidth,
    enlarge left by=0mm,
    before skip=5pt,
    after skip=5pt
  ]
  \normalsize
}{
  \end{tcolorbox}
}

\usepackage{amsmath}
\usepackage{pifont}
\usepackage{multirow}
\usepackage{hyperref}
\usepackage{subcaption} 
\usepackage[table]{xcolor}
\usepackage{algorithm}
\usepackage{algorithmicx}
\usepackage{algpseudocode}

\newcommand{\simworld}{\textsc{SimWorld}\xspace}

\newcommand{\websiteurl}{\href{https://simworld.org}{https://simworld.org}}

\definecolor{codegreen}{HTML}{478058}%
\definecolor{codegray}{rgb}{0.5,0.5,0.5}
\definecolor{codepurple}{HTML}{4F5E80} %
\definecolor{backcolour}{rgb}{0.95,0.95,0.92}

\lstdefinestyle{mystyle}{
    backgroundcolor=\color{backcolour},
    commentstyle=\color{codegreen},
    keywordstyle=\color{magenta},
    numberstyle=\tiny\color{codegray},
    stringstyle=\color{codepurple},
    basicstyle=\ttfamily\scriptsize,
    breakatwhitespace=false,
    breaklines=true,
    captionpos=b,
    keepspaces=true,
    frame=none,
    numbersep=5pt,
    showspaces=false,
    showstringspaces=false,
    showtabs=false,
    tabsize=2
}

\newtcolorbox{promptbox}[2][]{
    enhanced, 
    breakable,
    center title,
    left*=0pt, right*=0pt,
    boxsep=2pt, left=5pt, right=5pt,
    skin first=enhanced,
    skin middle=enhanced,
    skin last=enhanced,
    colback  = backcolour,
    fonttitle=\bfseries\rmfamily,
    fontupper=\scriptsize,
    title={\footnotesize\strut{#2}},
    #1
}

\tcbset{
  aibox/.style={
    width=\linewidth,
    top=10pt,
    bottom=4pt,
    before skip=2pt,
    colback=gray!5!white,        
    colframe=gray!100!white,      
    boxrule=0.4pt,               
    coltitle=black,              
    colbacktitle=white,          
    enhanced,
    center,
    attach boxed title to top left={yshift=-0.1in,xshift=0.15in},
    boxed title style={
      boxrule=0pt,               
      colback=white,             
      colframe=white,            
      fontupper=\small\bfseries, 
      left=4pt, right=4pt,       
    },
  }
}

\newtcolorbox{AIbox}[2][]{aibox,title=#2,#1}

\definecolor{kellygreen}{rgb}{0.3, 0.73, 0.09}
\definecolor{alizarin}{rgb}{0.82, 0.1, 0.26}
\newcommand{\cmark}{{\color{kellygreen} \ding{51}}}
\newcommand{\xmark}{{\color{alizarin} \ding{55}}}
\newcommand{\highlevel}{{\color{kellygreen} H}}
\newcommand{\lowlevel}{{\color{alizarin} L}}

\begin{document}

\input{configs/header}

\begin{center}
\maitrixtitle{
    \simworld: An Open-ended Realistic Simulator for \\ 
    Autonomous Agents in Physical and Social Worlds 
}

\setsymbol{colead}{*}
\setsymbol{coadvise}{$\dagger$}

\begin{authorlist}
\addauthor{Jiawei Ren}{ucsd,colead}
\addauthor{Yan Zhuang}{uva,colead} 
\addauthor{Xiaokang Ye}{ucsd,colead}
\addauthor{Lingjun Mao}{ucsd}
\addauthor{Xuhong He}{uiuc}
\addauthor{Jianzhi Shen}{jhu}
\addauthor{Mrinaal Dogra}{ucsd}
\addauthor{Yiming Liang}{purdue}
\addauthor{Ruixuan Zhang}{jhu}
\addauthor{Tianai Yue}{jhu}
\addauthor{Yiqing Yang}{polyu}
\addauthor{Eric Liu}{usc}
\addauthor{Ryan Wu}{jhu}
\addauthor{Kevin Benavente}{ucsd}
\addauthor{Rajiv Mandya Nagaraju}{usc}
\addauthor{Muhammad Faayez}{jhu}
\addauthor{Xiyan Zhang}{jhu}
\addauthor{Dhruv Vivek Sharma}{ucsd}
\addauthor{Xianrui Zhong}{uiuc}
\addauthor{Ziqiao Ma}{umich} 
\addauthor{Tianmin Shu}{jhu,coadvise}
\addauthor{Zhiting Hu}{ucsd,coadvise}
\addauthor{Lianhui Qin}{ucsd,coadvise}
\end{authorlist}


\affiliation{ucsd}{UCSD}
\affiliation{uva}{UVA}
\affiliation{jhu}{JHU}
\affiliation{umich}{UMich}
\affiliation{uiuc}{UIUC}
\affiliation{polyu}{PolyU}
\affiliation{purdue}{Purdue}
\affiliation{usc}{USC}

\printAffiliations{}

\printNotice{}

\vspace{-15pt}

\home \websiteurl

\addkeywords{Simulation, LLM Agents, ...}

\end{center}

\begin{figure}[!ht]
    \centering
    \vspace{-15pt}
    \includegraphics[width=0.98\linewidth]{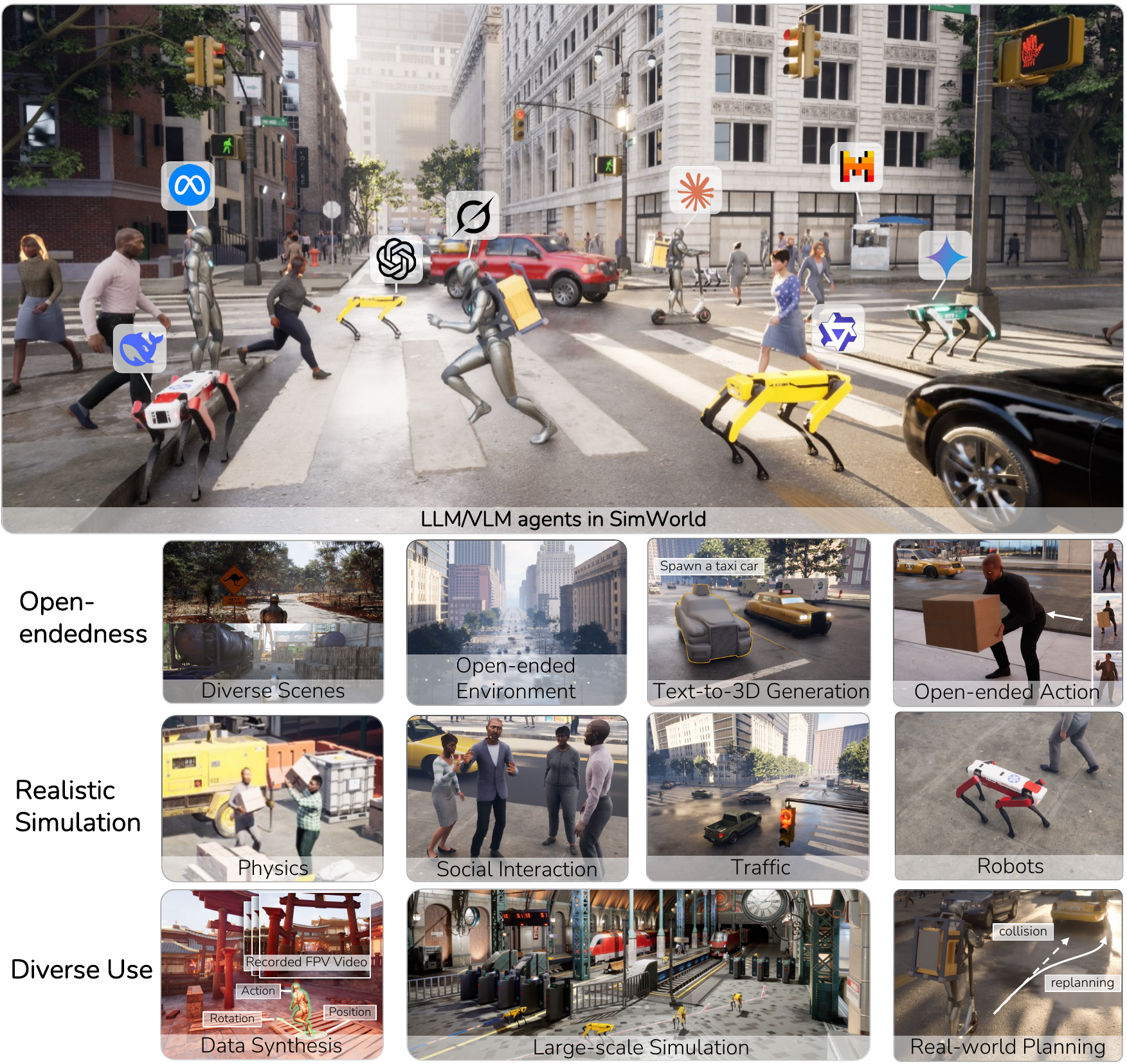}
    \vspace{-6pt}
    \caption{\textbf{An Overview of the \simworld Simulator}, featuring three key designs:
    (1) realistic, open-ended world simulation, (2) rich interface for LLM/VLM agents, and (3) diverse physical and social reasoning scenarios.
    }
    \label{fig:teaser}
\end{figure}
\begin{abstract}
While LLM/VLM-powered AI agents have advanced rapidly in math, coding, and computer use, their applications in complex physical and social environments remain challenging. Building agents that can survive and thrive in the real world (e.g., by autonomously earning income or running a business) requires massive-scale interaction, reasoning, training, and evaluation across diverse embodied scenarios. However, existing world simulators for such development fall short: they often rely on limited hand-crafted environments, simulate simplified game-like physics and social rules, and lack native support for LLM/VLM agents. 
We introduce \simworld, a new simulator built on Unreal Engine 5, designed for developing and evaluating LLM/VLM agents in rich, real-world-like settings. \simworld offers three core capabilities: {\it (1) realistic, open-ended world simulation}, including accurate physical and social dynamics and language-driven procedural environment generation; {\it (2) rich interface for LLM/VLM agents}, with multi-modal world inputs/feedback and open-vocabulary action outputs at varying levels of abstraction; and {\it (3) diverse extensible physical and social reasoning scenarios} that are easily customizable by users. We demonstrate \simworld by deploying frontier LLM agents (e.g.,  \texttt{GPT-4o}, \texttt{Gemini-2.5-Flash}, \texttt{Claude-3.5}, and \texttt{DeepSeek-Prover-V2}) on long-horizon multi-agent delivery tasks involving strategic cooperation and competition. The results reveal distinct reasoning patterns and limitations across models. 
We open-source \simworld and hope it becomes a foundational platform for advancing real-world agent intelligence across disciplines: \websiteurl.
\end{abstract}

\startcontents[main]
\vspace*{-5pt}
\section*{\textbf{Table of Contents}} 
\vspace*{-5pt}
{\setlength{\baselineskip}{0.9\baselineskip}
\printcontents[main]{}{1}{\setcounter{tocdepth}{3}}}
\vspace*{-10pt}
\clearpage

\clearpage

\section{Introduction}

Large language and vision models (e.g., LLMs and VLMs) have emerged as powerful foundations for building intelligent agents, demonstrating remarkable reasoning capabilities, particularly in structured domains such as mathematics, coding, and computer use (e.g., web browsing). However, these {\it mathematical} and {\it digital} settings are relatively clean, with well-defined rules and clear feedback. In contrast, the embodied {\it physical} and {\it social} worlds, where real-world agents and robots are ultimately expected to operate, are inherently complex, noisy, dynamic, and unpredictable. In such environments, agents must interact with rich and evolving contexts, from navigating urban spaces and interacting with humans, to pursuing long-term goals such as earning a living, building a career, or running an organization \cite{brohan2023can, driess2023palm, wang2023voyager}.

To advance embodied agent development, recent efforts have explored simulation environments that offer different interactive experiences for training and evaluation (Table~\ref{tab:comparison-table}). 
However, game-like platforms such as Minecraft~\cite{fan2022minedojo, white2025collaborating, wang2023voyager, long2024teamcraft, liu2024odyssey, Li_2025_CVPR} and Pokémon~\cite{ha2025gemini, anthropic2025visible} provide accessible setups for embodied interaction but lack realistic physical dynamics and social structures, limiting real-world generalization. Domain-specific simulators such as CARLA~\cite{2017carla} and AI2-THOR~\cite{kolve2017ai2}
target areas like autonomous driving and household robotics but are limited to narrow task scopes or static environments. Social sandboxes such as Virtual Village~\cite{park2023generative} and Project Sid \cite{al2024projectsid}
simulate interpersonal interactions in scripted, small-scale communities, but lack the open-endedness and scalability required for modeling richer social complexity. Moreover, many of these environments do not support natural language interfaces for goal setting, planning, and control, limiting their compatibility with modern LLM-based agents. 

To meet these growing demands, we present \simworld, a platform designed to support the development and evaluation of autonomous agents in complex, dynamic, and interactive environments. \simworld is grounded in three core design principles (Figure~\ref{fig:teaser}):

1) \textbf{Realistic, Open-Ended World Simulation.} \simworld advances simulation by integrating two key aspects:  {\it realistic} physical and social dynamics, and {\it open-ended, language-steerable} world generation. 
On the realism side, \simworld produces complex, dynamic environments grounded in physical laws (e.g., gravity, momentum) and enriched with dynamic elements such as lighting, weather, and pedestrian flow in city-scale 3D scenes. It also embeds socially grounded behaviors, such as obeying traffic signals and maintaining personal space, directly into agent logic to support realistic interactions. 
On the open-ended side, \simworld offers a broad range of scenes (e.g., city, countryside, wilderness, islands) and supports infinite environment expansion through procedural generation, including diverse road networks, building layouts, and urban configurations. Moreover, users or AI agents can modify scenes on-the-fly via natural language prompts (e.g., ``\texttt{add a tree next to the hospital}'').
Powered by \simworld's LLM-based editing and asset-generation modules, this capability enables adaptive, interactive world creation. 

2) \textbf{Rich Interface for LLM/VLM Agents.}  
\simworld provides a {\it Gym-like interface} that enables LLM/VLM agents to interact with simulated worlds using {\it open-ended natural language actions}. 
Agents can perceive rich multimodal observations (e.g., visual scenes, abstract layouts, and action feedback) and respond with high-level language commands. For example, an agent may reason and generate an abstract action, ``\texttt{sit on the nearest chair,}'' which \simworld automatically decomposes into a sequence of low-level actions (e.g., navigating through waypoints, sitting down). After executing the actions, the simulator provides updated observations and feedback, allowing the agent to refine its strategy and continue reasoning. This closed-loop interaction supports open-ended, language-driven behaviors and empowers agents to perform long-horizon reasoning at a proper abstraction level.

3) \textbf{Diverse Physical and Social Reasoning Scenarios.} Building on the above physically and socially grounded environments and the agentic interface, \textsc{SimWorld} naturally supports systematic evaluation and training of agent reasoning in diverse realistic, long-horizon settings. Beyond short, task-oriented behaviors, agents can pursue extended objectives such as earning money, developing a career trajectory, or running a multi-agent business, where strategic decisions compound over time and social dynamics influence outcomes. 
To illustrate how these capabilities integrate in practice, we showcase a \textit{Delivery Task}, a case study demonstrating how physical and social reasoning jointly lead to multi-agent collaboration and competition in \textsc{SimWorld}. The task models an urban delivery economy in which agents bid, invest, and share orders while navigating dynamic environments. With different personas, budgets, and tools (e.g., vehicles), agents develop diverse strategies shaped by their goals and changing conditions (e.g., fluctuating prices). The task highlights complex decision-making and long-horizon planning, where cooperation, competition, and emergent social behaviors arise naturally.


We deploy frontier LLMs as agents such as \texttt{GPT-4o}, \texttt{Claude-3.5-Sonnet}, \texttt{Gemini-2.5-Flash}, and others on the \textit{Delivery Task}. We observe that \texttt{Claude-3.5-Sonnet} and \texttt{DeepSeek-V3} earn the highest profits, but often behave erratically, such as overbidding on low-value orders or spending all their money on scooters they never use. In contrast, \texttt{Gemini-2.5-Flash} and \texttt{DeepSeek-Prover-V2} follow more conservative, stable strategies, trading peak performance for consistency.  Personality traits also shape agent behavior: conscientious agents focus on task completion, while open agents explore but frequently lose money. These findings expose both the strengths and limitations of LLM-based agents, while revealing rich, often unexpected behaviors that emerge from their interaction with complex environments. 

We open-source \simworld with the aim of establishing a foundational infrastructure for real-world agent research across disciplines. By supporting advanced LLM/VLM-based agents and enabling large-scale, realistic agent–environment and agent–agent interactions, \simworld expands the capabilities of modern {\it agent-based simulation (ABS)}. This allows researchers in robotics, business, public health, social science, education, and beyond to study complex systems and emergent behaviors in rich, dynamic, and controllable environments. More details of the \simworld project are available at \websiteurl.

\begin{table}[t]
\small
  \centering
  \vspace{-5pt}
  \caption{
  \textbf{Comparison of \simworld and Existing Simulators} across key dimensions: \textbf{Open-ended World} (procedural scene/asset generation, language-controllable editing), \textbf{Physical/Social Realism} (fidelity to real-world mechanics),
    \textbf{Action Space} (action abstraction level, open-vocabulary action space), \textbf{Agent Type} (types of controllable agents: Humanoid (Hum.), Robot, Drone or Vehicle (Veh.)), and \textbf{Physics Engine} (underlying simulation engine). 
    \highlevel\ means high-level actions (e.g., ``\texttt{deliver}'', ``\texttt{navigate to}''), and  
    \lowlevel\ means low-level actions (e.g., ``\texttt{forward by 1 step}''). 
    }
  \label{tab:comparison-table}
  \resizebox{\textwidth}{!}{%
    \begin{tabular}{@{}l cc c cc l l@{}}
      \toprule
      \multirow{2}{*}{\textbf{Simulator}}
        & \multicolumn{2}{c}{\textbf{Open-ended World}}
        & \multirow{2}{*}{\parbox[c]{2.7cm}{\centering\textbf{Physical/Social}\\\textbf{Realism}}}
        & \multicolumn{2}{c}{\textbf{Action Space}}
        & \multirow{2}{*}{\textbf{Agent Type}}
        & \multirow{2}{*}{\textbf{Physics Engine}} \\
      \cmidrule(r){2-3} \cmidrule(r){5-6}
        & Procedural  & Lang.-Ctrl 
        & 
        & Abstr. & Open-Vocab 
        & 
        & \\
      \midrule
      Minedojo~\cite{fan2022minedojo}          & \cmark        & \xmark  &  +  & \lowlevel        & \xmark    & Hum.                  & Minecraft             \\
      Mindcraft~\cite{white2025collaborating}  & \cmark       & \xmark &  +  & \highlevel        & \xmark    & Hum.               & Minecraft             \\
      MetaUrban~\cite{wu2025metaurban}         & \cmark          & \xmark & ++  & \lowlevel        & \xmark    & Veh.                & PyBullet              \\
      EmbodiedCity~\cite{gao2024embodiedcity}  & \xmark          & \xmark &  +++ & \lowlevel        & \xmark    & Drone/Veh.                     & Unreal Engine         \\
      CARLA~\cite{2017carla}                   & \xmark         & \xmark &  +++  & \lowlevel        & \xmark    & Veh.                    & UE \& Unity           \\
      GRUtopia~\cite{wang2024grutopia}         & \xmark          & \xmark &   ++  & \lowlevel        & \xmark    & Hum./Robot               & Isaac Sim             \\
      OmniGibson~\cite{li2024behavior}         & \xmark          & \xmark &   ++  &  \highlevel/\lowlevel        & \xmark  & Robot                  & Omniverse             \\
      AI2-THOR~\cite{kolve2017ai2}             & \cmark        & \xmark &  ++  &  \lowlevel        & \xmark    & Robot                     & Unity                 \\
      Habitat 3.0~\cite{puig2023habitat}       & \xmark           & \xmark &   ++  & \lowlevel        & \xmark   & Hum./Robot                  & Bullet                \\
      Genesis ~\cite{Genesis}                  & \cmark           & \xmark & +++    & \lowlevel      & \xmark    & Robot                    &          Taichi \\
      VirtualCommunity ~\cite{zhou2025virtual}& \cmark            & \xmark & ++    & \lowlevel       & \xmark   & Hum./Robot               & Genesis   \\
      UnrealZoo~\cite{zhong2025unrealzoo}& \xmark & \xmark & +++ & \lowlevel & \xmark & Hum./Robot/Veh. &Unreal Engine \\
      \midrule
      \textbf{\simworld}                & \cmark       & \cmark &  +++  &  \highlevel/\lowlevel      & \cmark    & Hum./Robot/Veh.       & Unreal Engine         \\
      \bottomrule
    \end{tabular}
  }
\end{table}

\begin{figure}[!ht]
    \centering
    \includegraphics[width=\linewidth]{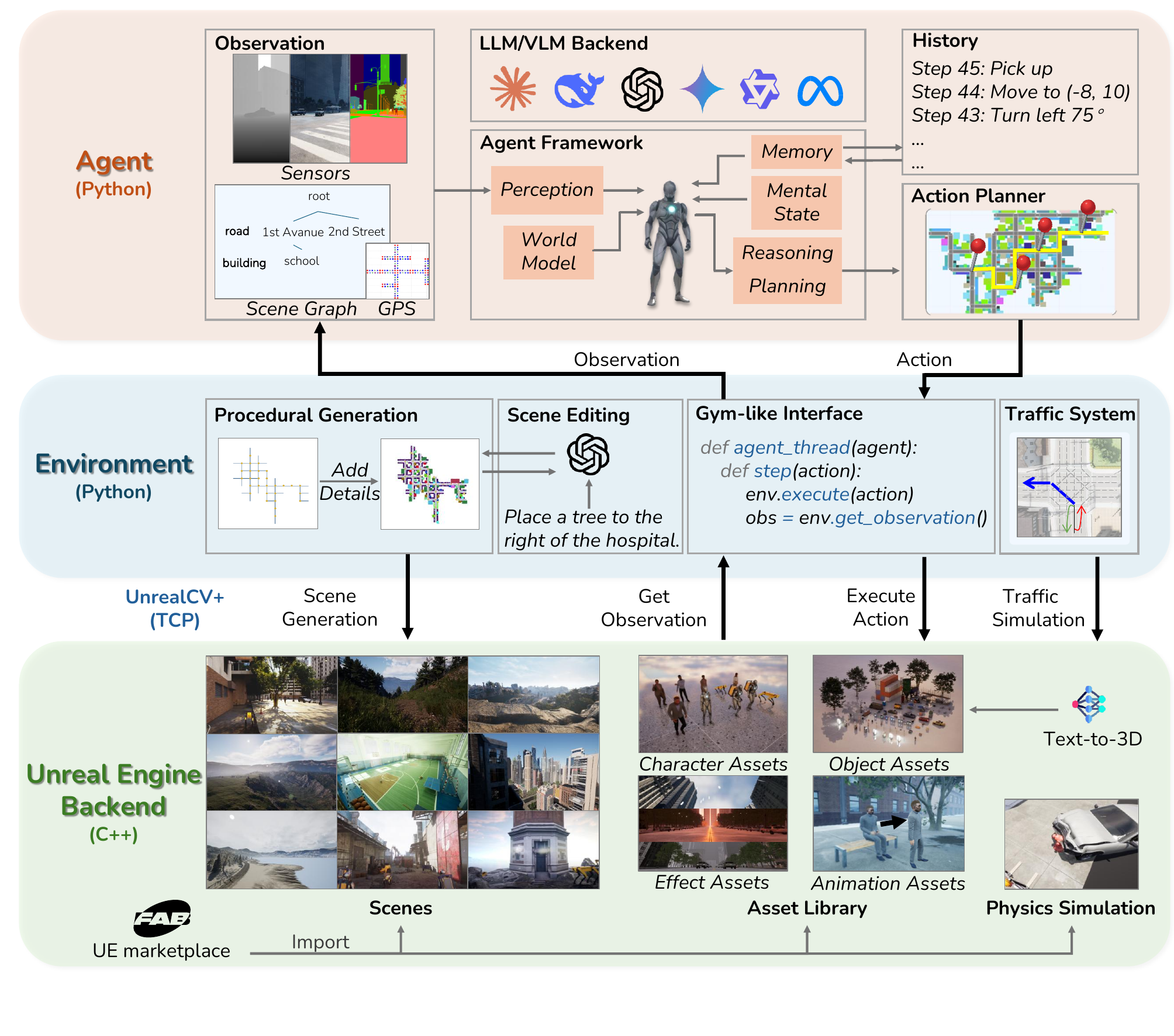}
    \vspace{-40pt}
    \caption{\textbf{Architecture of \simworld.} \simworld adopts a hierarchical, closed-loop architecture that decouples agent reasoning from high-performance rendering while maintaining coherent information flow across modules. At its core, the \textit{Unreal Engine Backend} provides high-fidelity scenes, assets, and physics, serving as the foundation for realistic simulation. Built upon it, the \textit{Environment} layer functions as an intermediary that abstracts the underlying rendering and physics into structured representations. It enables procedural city generation, traffic simulation, and exposes a Gym-like interface for agent interaction through \textit{UnrealCV+}. The \textit{Agent} layer operates on this interface, integrating LLM/VLM agents that interpret observations from the \textit{Environment}, perform reasoning, and issue actions that are subsequently executed through the \textit{Environment}’s connection to the \textit{Unreal Engine Backend}, thereby forming a closed perception–planning-action loop.}
    \label{fig:overview}
\end{figure}

\section{The \simworld Simulator}

Realistic, open-ended, and natively LLM/VLM-compatible simulators are crucial for advancing agent development in complex physical and social scenarios. \simworld takes a step toward this goal through a three-tier architecture as illustrated in Figure~\ref{fig:overview}. 
It separates the \textit{Unreal Engine Backend} (\S\ref{sec:ue_backend}) from two added Python layers: the \textit{Environment} layer providing infinite environment generation and standard Gym-like environment-agent interface (\S\ref{sec:environment_layer}), and the \textit{Agent} layer supporting diverse input observations, open-ended output actions, and different reasoning/planning components (\S\ref{sec:agent layer}). In addition, the \textit{UnrealCV+} communication module enables seamless interaction between the Unreal Engine backend and the Environment layer (\S\ref{sec:unrealcv}).

\subsection{Unreal Engine Backend}
\label{sec:ue_backend}
The Unreal Engine backend forms the foundation of \simworld, providing high-fidelity rendering and physics simulation. It consists of three tightly coupled modules: (1) \textbf{Scenes} (\S\ref{sec:environments}) supporting both procedurally generated and curated maps; (2) \textbf{Asset Library} (\S\ref{sec:rich_assets}) ensuring diverse and physically grounded content; and (3) \textbf{Physics Simulation} (\S\ref{sec:rich_assets}) governing realistic physical behaviors.

\subsubsection{Diverse Scenes}
\label{sec:environments}
\begin{figure}[t]
    \centering
    \includegraphics[width=\linewidth]{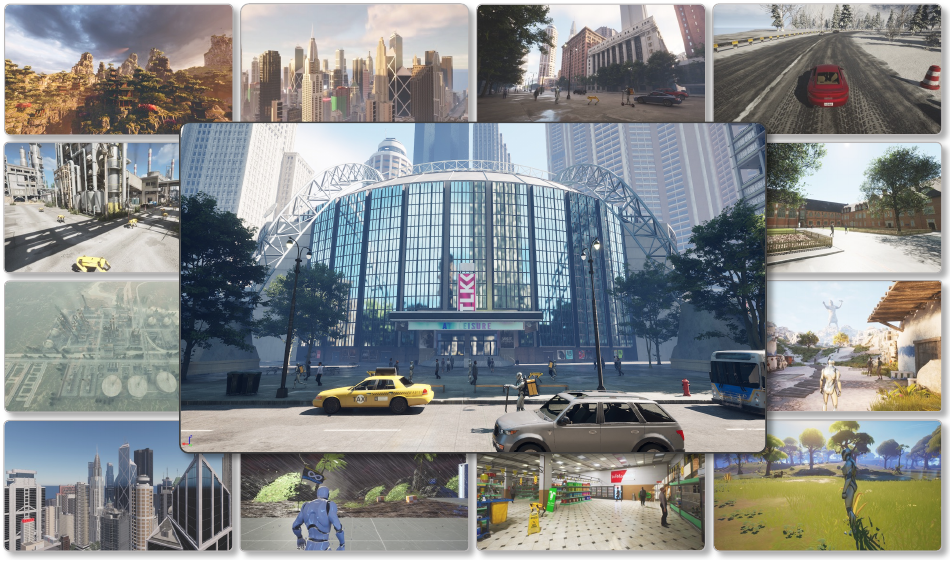}
    \vspace{-15px}
    \caption{\textbf{Example Scenes in \simworld.}}
    \label{fig:maps}
\end{figure}

\simworld supports two scene-building modes: handcrafted scenes and procedurally generated scenes.

\paragraph{Handcrafted Scenes.}
Thanks to \simworld’s foundation in Unreal Engine, users can easily import a large collection of high-quality environments directly from the Unreal Engine Marketplace\footnote[1]{\url{https://www.fab.com/}}
 or create custom scenes by hand. In our current implementation, we curate over 100 handcrafted scenes\footnote{The \simworld release includes all scenes as executable builds, available at \url{https://github.com/SimWorld-AI/SimWorld}.}
 spanning a wide variety of visual and structural styles, from ancient towns and natural landscapes to futuristic cities and imaginative fantasy worlds. Each scene provides distinct visual cues, spatial layouts, and interaction affordances, enabling thorough evaluation of embodied agents across diverse settings. Figure~\ref{fig:maps} illustrates several examples.

\paragraph{Procedurally Generated Scenes.}
Complementing these handcrafted assets, \simworld features a procedural generation module for automatically constructing diverse urban environments. Users can specify high-level parameters (e.g., city size, road density, layout style), and the system generates large numbers of city variants efficiently and consistently. This supports scalable experimentation under controlled, customizable conditions. Additional details are provided in Section~\ref{sec:pcg}.

\medskip

By combining high-fidelity handcrafted scenes with flexible procedural generation, \simworld offers a broad and extensible set of environments suitable for both controlled experiments and open-ended agent research.

\subsubsection{Rich Assets and Physics Realism}
\label{sec:rich_assets}
\simworld provides a comprehensive asset library to support realistic, physics-driven simulations across diverse environments.
The system integrates static assets (e.g., buildings) and dynamic assets (e.g., pedestrians), and further incorporates environmental factors (e.g., lighting, weather) to create immersive virtual worlds. It also supports a wide range of animations and interactions, enabling agents to perform diverse actions faithfully within the environments.

\paragraph{Object Assets.}
The object asset library forms the structural backbone of \simworld environments, where each scene can be viewed as a composition of multiple object assets. These assets include detailed material definitions and collision meshes, enabling a wide range of physically accurate interactions such as reflection, occlusion, and contact dynamics. Overall, they can be broadly grouped into three categories:
(i) \textit{Building assets:} Primary structural elements of urban scenes, covering a wide range of architectural types (e.g., residential, commercial, industrial) and supporting both indoor and outdoor environment construction. 
(ii) \textit{Vegetation assets:} Natural elements such as trees, grass, and shrubs, modeled with realistic material appearance and optional seasonal variations. 
(iii) \textit{Urban prop assets:} Fine-grained objects such as benches, mailboxes, lampposts, and traffic signs, enabling diverse agent interactions such as sitting, opening, or manipulating objects. 


\paragraph{Text-to-3D Asset Generation.}
To further expand the range of available objects, \simworld introduces an \textit{Asset Generation Pipeline} based on recent Text-to-3D models~\cite{hunyuan3d22025tencent}.
This system allows users to describe assets in natural language, automatically generating 3D objects with consistent scale, texture, and physical properties. The generated assets can be seamlessly integrated into the simulator, inheriting various properties (e.g., materials, lighting, collision configurations) compatible with UE’s physics engine.

\begin{figure}
    \centering
    \includegraphics[width=\linewidth]{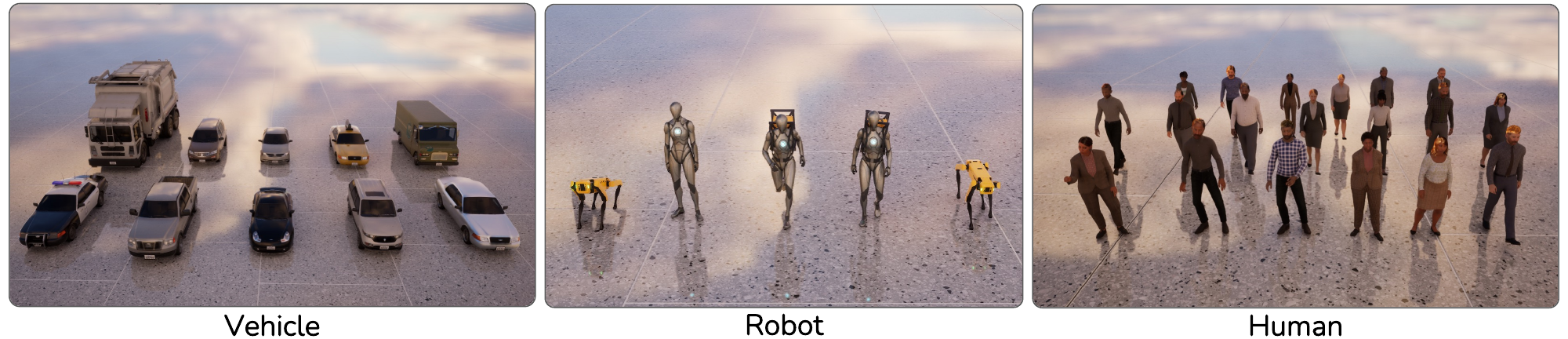}
    \caption{ \textbf{Embodied Agents. }\simworld supports three types of agent embodiments: vehicle, robot, and human.}
    \label{fig:embodiment}
\end{figure}

\paragraph{Characters and Embodiments.}
Character assets in \simworld represent embodied entities capable of acting, navigating, and interacting within the virtual environment. The system supports three primary types of agent embodiments: human, vehicle, and robot (Figure~\ref{fig:embodiment}). Human embodiments capture diverse human appearances and employ fully rigged skeletal structures that enable realistic animations produced through coordinated bone articulation, such as running or carrying objects. Vehicle embodiments reproduce a range of real-world transportation modes (e.g., buses, cars) and implement accurate physical driving dynamics, such as acceleration, steering, braking, and traction. Robotic embodiments model specific categories of robots (e.g., quadruped systems) with realistic actuation, joint control, and sensing modules, making them suitable for evaluating robot locomotion and stability across different environments and tasks. All these embodiments operate within a unified physics framework and share common attributes (e.g., mass, inertia, contact forces), which ensures consistent handling of physical properties and interactions across all entities.

\paragraph{Weather and Lighting.}
\simworld supports a wide range of lighting and weather conditions. The lighting system models multiple light types (e.g., directional, ambient, and dynamic sources) with controllable parameters such as intensity, orientation, and color temperature. The weather system supports a variety of conditions (e.g., rain, snow, and fog) that influence visual appearance and drive atmospheric effects, including phenomena like fog-induced light scattering. Together, these components recreate the complexity and dynamism of real-world environments, enabling the study of embodied agents’ perception and adaptation under realistic conditions.

\paragraph{Physical Dynamics and Animations.}
Powered by Unreal Engine, \simworld provides accurate and continuous physical simulation. Unlike popular agent environments such as Minecraft~\cite{fan2022minedojo, yu2024mineland}, which rely on discrete, block-based mechanics without real gravity or inertia, \simworld models real-world physical dynamics. Agents are subject to physical forces that produce grounded behaviors like sliding down slopes or tripping over steps. These effects produce physically grounded, embodied interactions. By combining Unreal Engine's physics engine with physically informed animations (e.g., motion blending, inverse kinematics, collision responses), \simworld maintains coherence between motion and environmental forces, enabling believable and adaptable agent behaviors in complex environments.




\subsection{Environment}
\label{sec:environment_layer}

\simworld introduces an environment layer on top of the Unreal Engine backend (Figure~\ref{fig:overview}). This layer manages the creation and organization of simulated environments and provides a clean abstraction that enables easy deployment of agents into Unreal Engine–based worlds through AI-native, user-friendly interfaces, without requiring users to handle the complexities of the underlying UE system. 
Specifically, the environment layer integrates modules for Procedural City Generation (\S\ref{sec:pcg}), LLM-based Scene Editing (\S\ref{sec:scene_editing}), Traffic Systems (\S\ref{sec:traffic_system}), and, crucially, a Gym-like Interface (\S\ref{sec:simulation_modes}) for agent–environment interaction. It also offers an auxiliary Waypoint System (\S\ref{sec:waypoint_system}) that simplifies agent navigation within complex worlds.

\subsubsection{Procedural City Generation} 
\label{sec:pcg}
\begin{algorithm}[!t]
\caption{Procedural City Layout Generation using QuadTree}
\label{algo1}
\begin{algorithmic}[1]
\State \textbf{Input:} Configuration parameters $config$
\State \textbf{Output:} Final QuadTree $Q_{city}$ representing the city layout
\State Initialize empty QuadTree $Q_{city}$

\noindent\hspace{-1.7em}\colorbox{blue!5}{%
\begin{minipage}{\dimexpr\textwidth-2\fboxsep}
\begin{algorithmic}[1]
\setcounter{ALG@line}{3}
\If{$s = \text{road}$}
    \State Generate road network via growth-based model \Comment{Procedural street expansion}
    \State Insert road geometries into $Q_s$ \Comment{Store road segments}
    \State \textbf{Merge} $Q_s$ \textbf{into} $Q_{city}$ \Comment{Integrate road layout}
\EndIf
\end{algorithmic}
\end{minipage}}

\noindent\hspace{-1.7em}\colorbox{cyan!5}{%
\begin{minipage}{\dimexpr\textwidth-2\fboxsep}
\begin{algorithmic}[1]
\setcounter{ALG@line}{7}
\If{$s = \text{building}$}
    \State Sample building candidates (orientation, position)
    \State Reject invalid samples by collision test \Comment{Spatial consistency filtering}
    \State Greedy fill remaining gaps with valid buildings
    \State Insert buildings into $Q_s$
    \State \textbf{Merge} $Q_s$ \textbf{into} $Q_{city}$ \Comment{Integrate building layout}
\EndIf
\end{algorithmic}
\end{minipage}}

\noindent\hspace{-1.7em}\colorbox{teal!5}{%
\begin{minipage}{\dimexpr\textwidth-2\fboxsep}
\begin{algorithmic}[1]
\setcounter{ALG@line}{13}
\If{$s = \text{street element}$}
    \State Sample decorative/environmental elements
    \State Reject overlapping samples by collision test
    \State Insert detail elements into $Q_s$
    \State \textbf{Merge} $Q_s$ \textbf{into} $Q_{city}$ \Comment{Integrate street-level details}
\EndIf
\end{algorithmic}
\end{minipage}}

\State \Return $Q_{city}$
\end{algorithmic}
\end{algorithm}

\begin{figure}[t]
    \centering
    \begin{subfigure}{0.48\linewidth}
        \centering
        \includegraphics[width=\linewidth]{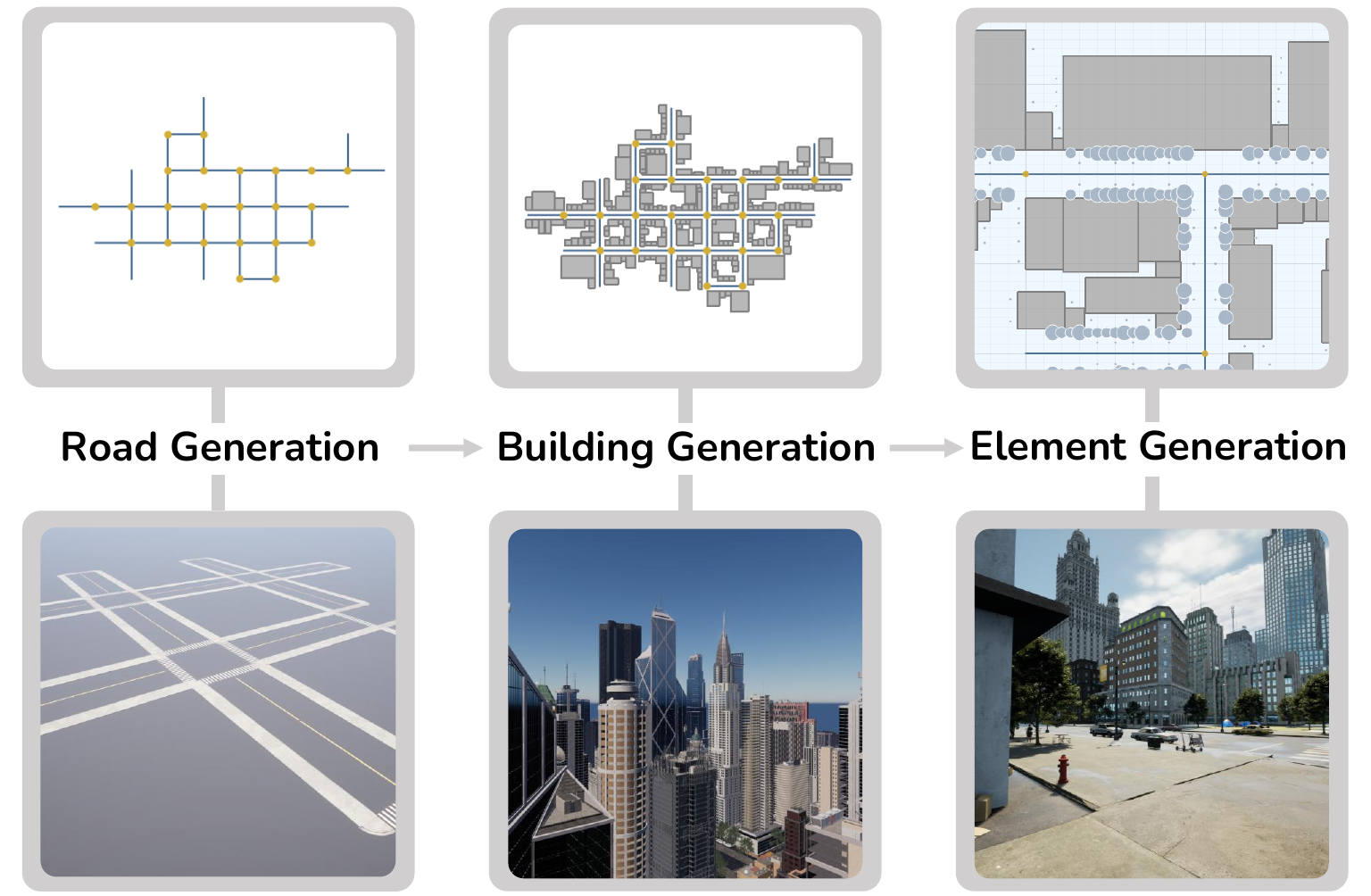}
        \caption{Procedural City Generation}
        \label{fig:citygen}
    \end{subfigure}
    \hfill
    \begin{subfigure}{0.47\linewidth}
        \centering
        \includegraphics[width=\linewidth]{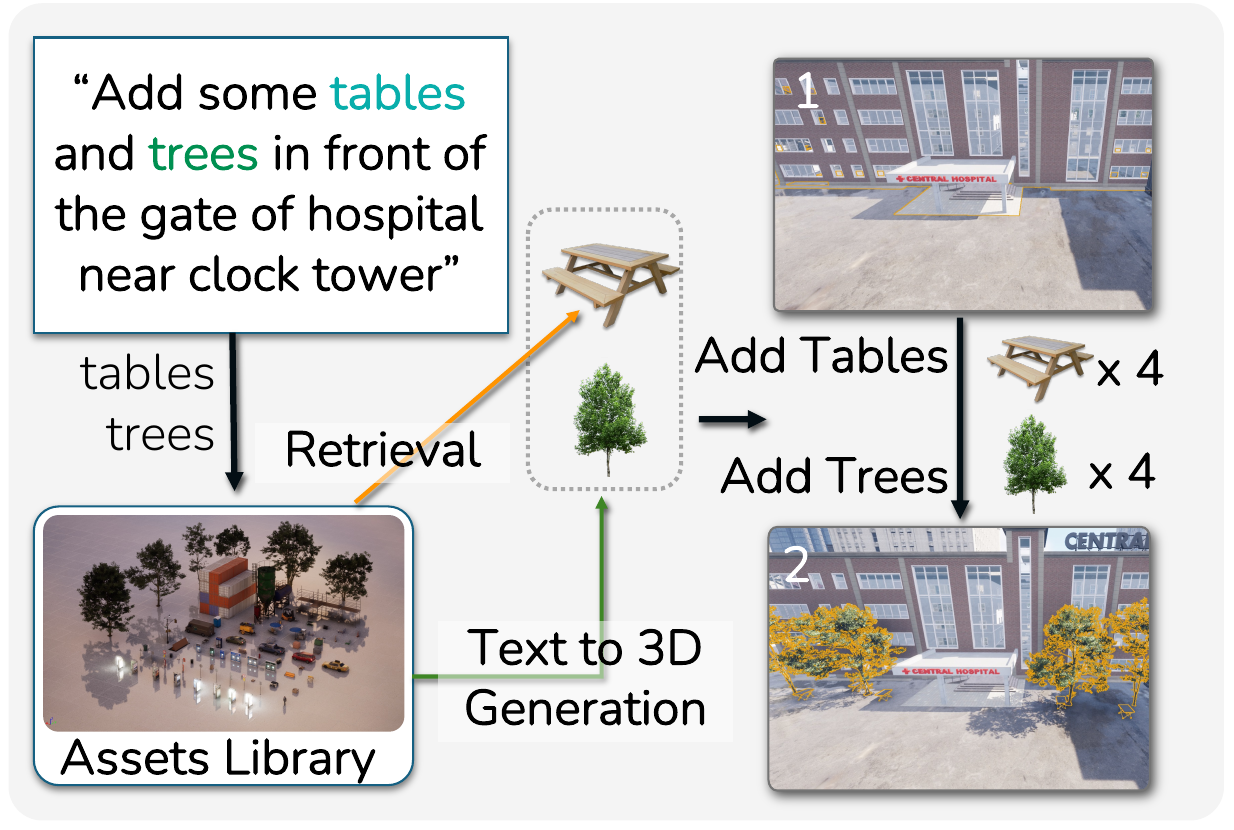}
        \caption{LLM-Controllable Scene Edit}
        \label{fig:LLM-based-Scene-Editing}
    \end{subfigure}

    \caption{\textbf{Overview of Procedural City Generation and LLM-Based Scene Editing.}}
    \label{fig:two-figs}
\end{figure}

Previous simulators typically rely on a limited set of hand-crafted scenes (e.g., 15 scenes in CARLA and 211 scenes in Habitat 3.0).
\simworld develops a procedural generation system (Figure~\ref{fig:citygen}) capable of producing diverse, unlimited urban environments, including road networks, building layouts, dynamic traffic, and fine-grained elements like street furniture, enabling effectively infinite simulation scenarios. All parameters (e.g., city size, building density, vehicle and pedestrian count) are customizable, allowing users to generate varied and controllable environments with minimal manual effort.

Inspired by \cite{phiresky2024procedural}, \simworld’s procedural generation system adopts a modular and extensible architecture. The pipeline proceeds through three sequential stages: road generation, building generation, and street element generation, each progressively enriching the environment with structural and visual complexity. The system constructs a hierarchical scene graph based on a quadtree data structure as illustrated in Algorithm \ref{algo1}.

\paragraph{Road Generation.} Road generation defines the structural backbone of the city layout. Roads are first initialized and then expanded using a spanning-tree–based algorithm with a priority queue that balances depth and branching during network construction. Additional procedures, including road-end attachment and intersection validation, maintain topological coherence and realism in the generated layout.

\paragraph{Building Generation.}Following the road generation stage, the pipeline proceeds to building generation, where building assets are procedurally instantiated along road segments. Candidate locations are sampled and validated for spatial feasibility to prevent overlap. A greedy placement strategy then fills residual gaps near intersections and road ends, improving spatial utilization and maintaining visual consistency.

\paragraph{Street Element Generation.}Finally, street element generation adds detailed environmental elements (e.g., trees, road cones, benches, and parked vehicles). Elements are categorized and positioned based on contextual zones, either adjacent to buildings or along sidewalks. While strict collision enforcement is relaxed to maintain performance, placement still respects basic accessibility and spatial coherence constraints.



\subsubsection{LLM-based Scene Editing}
\label{sec:scene_editing}


Beyond procedural generation, \simworld supports natural language-based scene editing (Figure~\ref{fig:LLM-based-Scene-Editing}), enabling dynamic world construction through open-ended instructions. Users or AI agents can modify scenes on-the-fly with commands such as “\texttt{add a red sports car next to the hospital near a museum}”. \simworld contains a retrieval-augmented LLM-based scene agent that grounds the command by querying the current environment's scene graph. The agent identifies the intended location using spatial anchors (e.g., “\texttt{hospital}”) and contextual landmarks (“\texttt{museum}”), retrieves a matching asset from a library, and inserts it accordingly. If a suitable asset is unavailable, the agent invokes an off-the-shelf text-to-3D generation model \cite{hunyuan3d22025tencent} to synthesize a new object from the prompt (“\texttt{red sports car}”), converts it into a compatible format, and integrates it into the environment. This approach enables semantically grounded, spatially coherent, and scalable world construction, laying the foundation for interactive and compositional simulation.



\subsubsection{Waypoint System} 
\label{sec:waypoint_system}

\begin{figure}[t]
    \centering
    \includegraphics[width=\linewidth]{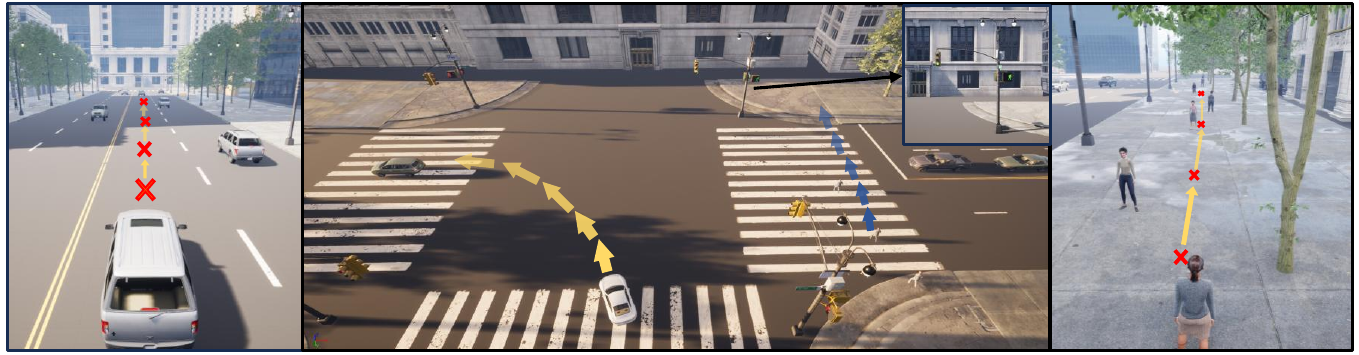}
    \vspace{-15px}
    \caption{\textbf{Overview of Waypoint System.} Vehicles and pedestrians navigate through the environment by following waypoints.}
    \label{fig:waypoint}
\end{figure}

\simworld implements a waypoint system that provides a structured representation of navigable space to support agent navigation and path planning (Figure~\ref{fig:waypoint}). As an auxiliary abstraction layer, the waypoint system simplifies movement by offering a clean, graph-based representation of where agents can go and how they can get there. It forms the spatial backbone for both the traffic system (\S\ref{sec:traffic_system}) and the action planner (\S\ref{sec:action_planner}), enabling agents to move efficiently through complex environments.

The system includes two complementary waypoint representations, coarse-grained and fine-grained, which together create a unified navigation graph. Coarse waypoints capture high-level connectivity (e.g., roads, intersections), while fine-grained waypoints represent detailed walkable paths. This hierarchical structure enables flexible and robust navigation behaviors, including lane following, turning, detouring, and obstacle avoidance.

\paragraph{Coarse-grained Waypoints.}
The coarse-grained waypoints are generated from the geometric outputs of the procedural city generation module (\S\ref{sec:pcg}), including road centerlines and intersection coordinates. These waypoints represent major structural points within the road network and capture the primary connectivity among different routes.

\paragraph{Fine-grained Waypoints.}
The fine-grained waypoints are interpolated along the roads between coarse-grained waypoints. These additional points increase the density of the navigation graph, allowing agents to follow smoother and more continuous trajectories. Parameters such as interpolation step size and spatial offset magnitude can be customized by users.

\begin{figure}[t]
    \centering
    \includegraphics[width=0.85\linewidth]{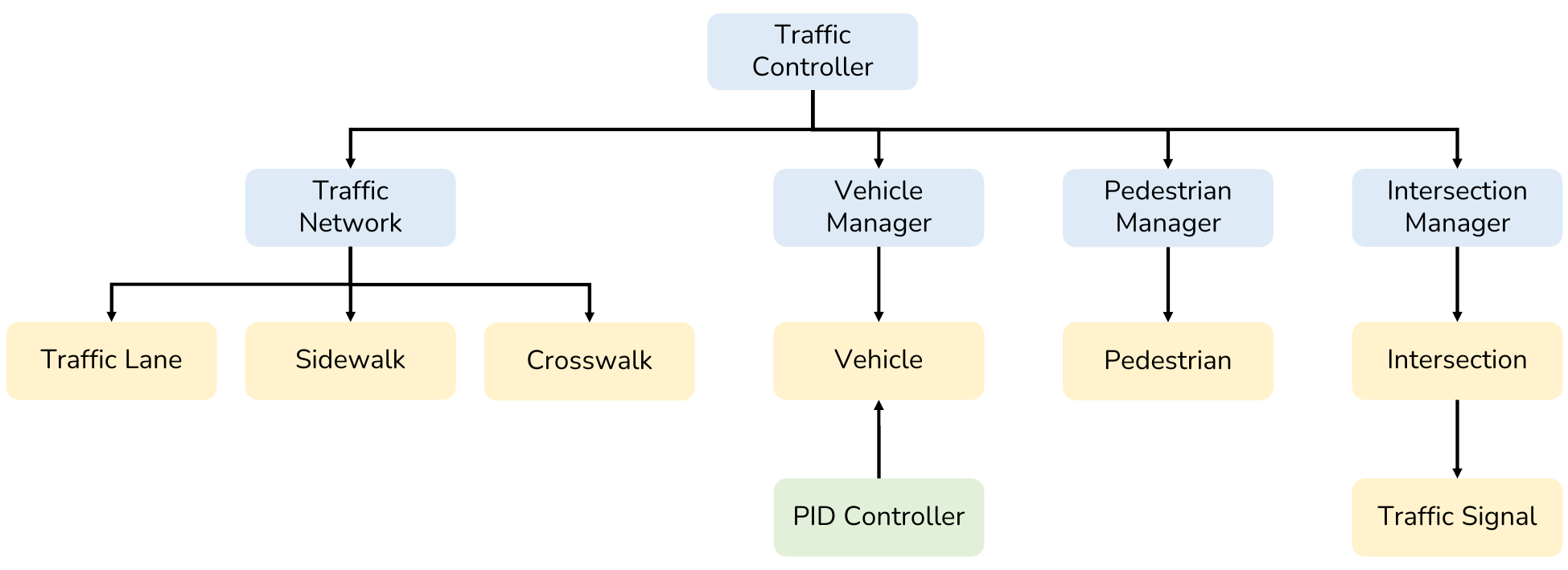}
    \vspace{-6pt}
    \caption{\textbf{Architecture of Traffic System in \simworld.}}
    \label{fig:traffic_arch}
\vspace{-10pt}
\end{figure}

\subsubsection{Traffic System} 
\label{sec:traffic_system}

The traffic system in \simworld simulates dynamic road usage involving both vehicles and pedestrians. It models realistic traffic flow through modules of vehicle spawning, route assignment, intersection control, and pedestrian movement (Figure~\ref{fig:traffic_arch}). By managing interactions among agents and coordinating traffic signals, the system supports complex urban phenomena such as congestion, pedestrian crossings, and traffic light synchronization.

The traffic simulation supports route assignment, intersection control, and pedestrian flow simulation, running on a fixed-timestep update loop for consistent and deterministic updates (Algorithm~\ref{alg:traffic_simulation}). Vehicle motion is governed by a proportional–integral–derivative (PID) controller, with empirically tuned parameters for realistic acceleration, braking, and turning dynamics \cite{jain2024comprehensive}. Pedestrian motion follows a lightweight model that adjusts pedestrians' orientations incrementally toward their goals based on angular differences. To simulate realistic patterns, \simworld uses a stochastic routing policy at intersections, i.e., agents select outgoing routes according to predefined probability distributions. This stochastic behavior introduces natural variability and enhances scene diversity.

The traffic system is built upon the waypoint system (\S\ref{sec:waypoint_system}), enabling traffic simulation that generalizes to any procedurally generated city layout. Using the waypoints, the system calculates the detailed traffic areas (e.g., road lanes, sidewalks, and crosswalks) and procedurally instantiates vehicles, pedestrians, and traffic signals accordingly. Three specialized managers coordinate these processes:

\paragraph{Vehicle Manager.} Vehicle manager initializes vehicles along designated traffic lanes and assigns either predefined or dynamically generated routes through the navigation network.

\paragraph{Pedestrian Manager.} Pedestrian manager spawns pedestrians on sidewalks and governs their motion patterns, including crossing behavior and local avoidance at intersections.

\paragraph{Intersection Manager.} Intersection manager detects intersections within the traffic network and deploys traffic signals that regulate right-of-way according to configurable timing cycles or adaptive control policies.

Together, these components constitute a unified traffic simulation pipeline, enabling the virtual city to exhibit realistic, adaptive, and scalable mobility dynamics across diverse urban layouts.

\begin{algorithm}[t]
\caption{Simulation Loop for Urban Traffic Environments}
\label{alg:traffic_simulation}
\begin{algorithmic}[1]
\State \textbf{Initialize:} Sample initial states for vehicles $\mathcal{V}$, pedestrians $\mathcal{P}$, and traffic signals $\mathcal{S}$.
\State Set simulation time $t \leftarrow 0$.
\While{$t < T_{max}$} \Comment{Main simulation loop}
    \State \colorbox{blue!5}{\Call{UpdateVehicles}{$\mathcal{V}, \mathcal{P}, \mathcal{S}$}}
    \State \colorbox{cyan!5}{\Call{UpdatePedestrians}{$\mathcal{P}, \mathcal{V}, \mathcal{S}$}}
    \State \colorbox{teal!5}{\Call{UpdateSignals}{$\mathcal{S}, t$}}
    \State $t \leftarrow t + \Delta t$
\EndWhile

\noindent\hspace{-1.7em}\colorbox{blue!5}{%
\begin{minipage}{\dimexpr\textwidth-2\fboxsep}
\begin{algorithmic}[1]
\setcounter{ALG@line}{9}
\Function{UpdateVehicles}{$\mathcal{V}, \mathcal{P}, \mathcal{S}$}
    \ForAll{$v \in \mathcal{V}$}
        \State Perceive environment $(\mathcal{V}, \mathcal{P}, \mathcal{S})$
        \State Execute driving model (throttle, brake, steering)
        \State Update $v$'s state (position, velocity)
    \EndFor
\EndFunction
\end{algorithmic}
\end{minipage}}

\noindent\hspace{-1.7em}\colorbox{cyan!5}{%
\begin{minipage}{\dimexpr\textwidth-2\fboxsep}
\begin{algorithmic}[1]
\setcounter{ALG@line}{15}
\Function{UpdatePedestrians}{$\mathcal{P}, \mathcal{V}, \mathcal{S}$}
    \ForAll{$p \in \mathcal{P}$}
        \State Perceive $(\mathcal{V}, \mathcal{S})$, execute walking logic
        \State Update $p$'s position
    \EndFor
\EndFunction
\end{algorithmic}
\end{minipage}}

\noindent\hspace{-1.7em}\colorbox{teal!5}{%
\begin{minipage}{\dimexpr\textwidth-2\fboxsep}
\begin{algorithmic}[1]
\setcounter{ALG@line}{20}
\Function{UpdateSignals}{$\mathcal{S}, t$}
    \ForAll{$s \in \mathcal{S}$}
        \State Update $s$ according to timing plan or adaptive policy
    \EndFor
\EndFunction
\end{algorithmic}
\end{minipage}}

\end{algorithmic}
\end{algorithm}

\subsubsection{Gym-like Interface for Agent-Environment Interaction}
\label{sec:simulation_modes}

\simworld provides a standard Gym-like interface, enabling seamless integration with existing reinforcement learning pipelines and agent frameworks. Because this interface follows the widely adopted API conventions of Gym \citep{gymnasium2023}, such as standardized \texttt{reset()}, \texttt{step()}, and observation–action exchange (Figure~\ref{fig:overview}), it becomes straightforward for users to plug in their RL agents and immediately begin interacting with \simworld’s simulated environments. This design significantly lowers the barrier for conducting large-scale experimentation, benchmarking, and agent–environment interaction studies using modern LLM/VLM or policy-based agents.

To support a broad variety of research goals ranging from open-ended simulations to highly controlled evaluations, \simworld offers two simulation modes inspired by prior work such as CARLA~\cite{2017carla}: asynchronous and synchronous execution.

\paragraph{Asynchronous Mode.}
In asynchronous mode, each agent runs in its own thread and advances independently, without waiting for other agents to finish their reasoning or action generation. Agents pull observations from a centralized buffer and submit actions whenever they are ready. The environment processes all received actions at fixed intervals (default: 0.1s), allowing real-time, continuous, and scalable multi-agent interactions. This mode is ideal for large-scale, open-ended, or exploratory simulations where throughput, diversity, and responsiveness are key.

\paragraph{Synchronous Mode.}
In synchronous mode, all agents advance in lockstep: the simulator proceeds to the next step only after every agent has submitted its action. This ensures strict temporal alignment between perception and control, making the mode particularly suitable for experiments requiring reproducibility, coordinated multi-agent behavior, or high-quality data collection (e.g., video generation or RL training with fixed step timing).

\subsection{Agent}
\label{sec:agent layer}
\simworld provides a unified interface for LLM/VLM agents, supporting a flexible \textbf{Agent Framework} (\S\ref{sec:agent_framework}), a diverse \textbf{Observation Space} (\S\ref{sec:observation_space}) and an open-ended \textbf{Action Space} (\S\ref{sec:action_space}). This interface is designed to accommodate both low-level control and high-level reasoning for LLM/VLM agents through an \textbf{Action Planner} (\S\ref{sec:action_planner}) module, serving as the primary entry point for agent–environment interaction.

\subsubsection{Agent Framework}
\label{sec:agent_framework}

The agent framework in \simworld defines a unified interface that structures the full agent loop of perception, reasoning, planning, and execution. Across different embodiments—humanoids, vehicles, and robots—the framework provides a common control pipeline.

Each agent first acquires observations from \simworld’s observation space  (\S\ref{sec:observation_space}) via API calls (e.g., \texttt{get\_camera\_observation()}, \texttt{get\_agent\_location()}). These observations (e.g., visual inputs, scene graphs) are then processed by the agent’s reasoning backend, which may incorporate LLMs, VLMs, VLAs, or other decision-making models. Based on these observations, agents can employ any advanced reasoning or planning algorithms \citep{guo2025deepseek,hao2023reasoning,al2024projectsid}.

Reasoning outputs may be expressed in natural language (e.g., “\texttt{sit on the nearest chair}”) or in structured formats such as function calls. Both formats are compatible with the action planner (\S\ref{sec:action_planner}), which interprets them into executable low-level actions.

The framework is also highly extensible. Researchers can plug in advanced reasoning components such as world models \citep{hu2023language,xing2025critiques}, memory systems \citep{ho2025arcmemo,wang2024memoryllm}, or mental-state modules (e.g., emotions or preferences), enabling broad investigation into long-horizon reasoning, planning, and embodied intelligence.


\subsubsection{Observation Space}
\label{sec:observation_space}

\simworld provides multiple observation modalities for agent perception and reasoning. The observation space is organized into two primary categories: visual observations and structured semantic information (Figure~\ref{fig:overview}).

\paragraph{Visual Observations.}
Agents can access three types of camera-based inputs from a first-person view: (1) color images capturing the raw visual appearance of the environment, (2) depth maps encoding geometric distance from the agent’s viewpoint, and (3) semantic segmentation masks providing pixel-level object category information. 

\paragraph{Structured Semantic Observations.}
Beyond pixel-based perception, \simworld exposes high-level spatial and semantic representations, including a semantic scene graph and GPS-like localization information. The scene graph encodes entities, attributes, and relational structures within the environment, offering a symbolic abstraction of the 3D world. The localization interface specifies each agent’s or object’s position and orientation, enabling precise reasoning about spatial relationships.


\subsubsection{Action Space} 
\label{sec:action_space}
\simworld enables open-vocabulary action execution by organizing the action space into two hierarchical layers: high-level semantic actions and low-level primitive actions:

\paragraph{High-Level Semantic Actions.} 
To facilitate abstract reasoning and long-horizon decision-making, agents can issue natural language commands. These commands are interpreted and executed by the built-in action planner (\S\ref{sec:action_planner}), enabling flexible, open-ended behaviors (e.g., “\texttt{sit on the nearest chair}”).

\paragraph{Low-Level Primitive Actions.} 
Primitive actions provide fine-grained control over agents. 
Vehicles support continuous control signals (e.g., ``\texttt{acceleration}'', ``\texttt{braking}'', and ``\texttt{steering}''). 
Robots allow continuous translation and rotation (e.g., ``\texttt{forward}'', ``\texttt{backward}'', ``\texttt{lateral\_movement}'', and ``\texttt{rotation}''). Human agents can navigate (``\texttt{move}'', ``\texttt{turn}'') and perform interactive actions, including human–object (e.g., ``\texttt{pick\_up}'', ``\texttt{drop}'', ``\texttt{sit}''), human–vehicle (e.g., ``\texttt{enter\_car}'', ``\texttt{exit\_car}''), and human–human (e.g., ``\texttt{wave\_hands}'', ``\texttt{discussion}'') interactions. A complete list of supported actions is provided in Table~\ref{tab:simworld-actionspace-full}. 

\begin{table}[t]
\caption{\textbf{Low-Level Primitive Actions in \simworld.}}
\label{tab:simworld-actionspace-full}
\centering
\footnotesize
\setlength{\tabcolsep}{8pt}
\renewcommand{\arraystretch}{1.1}

\begin{tabular}{lll}
\toprule
\textbf{Action} & \textbf{Agent Type} & \textbf{Description} \\
\midrule
\rowcolor{teal!10}
\multicolumn{3}{c}{\textit{Object Interaction Actions}} \\
\rowcolor{teal!2}
Pick Up / Drop Off & Humanoid & Grasp or release an object \\
\rowcolor{teal!2}
Carry / Put Down Heavy Object & Humanoid & Transport and place large objects \\
\rowcolor{teal!2}
Sit Down / Stand Up & Humanoid & Transition between seated and standing states \\
\rowcolor{teal!2}
Open Door / Enter / Exit Car & Humanoid & Interact with doors or vehicles \\
\rowcolor{teal!2}
Ride Scooter & Humanoid & Control and ride a scooter \\[2pt]

\rowcolor{green!8}
\multicolumn{3}{c}{\textit{Observation Actions}} \\
\rowcolor{green!1}
Look Up / Down & Humanoid, Dog & Adjust gaze vertically \\
\rowcolor{green!1}
Focus & Humanoid, Dog & Narrow or widen the field of view \\
\rowcolor{green!1}
Take Photo & All & Capture current view as an image \\[2pt]

\rowcolor{cyan!6}
\multicolumn{3}{c}{\textit{Social Actions}} \\
\rowcolor{cyan!1}
Have Conversation / Discuss & Humanoid & Exchange verbal communication \\
\rowcolor{cyan!1}
Point Direction / Wave Hand & Humanoid & Use gestures for social signaling \\
\rowcolor{cyan!1}
Argue with Body Language & Humanoid & Express disagreement through gestures \\

\rowcolor{blue!6}
\multicolumn{3}{c}{\textit{Navigation Actions}} \\
\rowcolor{blue!1}
Move Forward / Step Forward–Backward & Humanoid, Dog & Move or step in the current direction for a short duration \\
\rowcolor{blue!1}
Rotate / Steering & Humanoid, Dog, Vehicle & Adjust facing or steering direction \\
\rowcolor{blue!1}
Throttle / Brake & Vehicle & Accelerate or decelerate the vehicle \\
\rowcolor{blue!1}
Stop & All & Halt all current motion \\[2pt]

\bottomrule
\end{tabular}
\end{table}

\subsubsection{Action Planner}
\label{sec:action_planner}

\simworld includes an action planner module that bridges high-level reasoning with low-level execution, allowing researchers to focus on abstract planning without needing to manage embodiment-specific control details. The planner consists of two components: a {\it parser} and an {\it executor}. The parser receives high-level plans from the agent, often expressed in natural language or structured function calls, and translates them into sequences of low-level primitive actions. The executor then carries out these actions step by step, conditioned on the current environment state.

To support diverse research objectives, \simworld provides two executor variants: a {\it rule-based} executor, which operates on abstract city-layout information, and a {\it visual-based} executor, which directly consumes visual observations from the simulator. The latter enables seamless integration with VLMs or VLAs, supporting end-to-end perception–reasoning–action pipelines.

By handling the translation from high-level intent to low-level control, the action planner enables agents to perform long-horizon, semantic planning while \simworld automatically manages movement, navigation, and interaction details.

For example, when the action planner receives a plan such as ``\texttt{go to the nearest chair and sit down}'', the parser first decomposes the instruction into an action list: ``\texttt{navigate}'' and ``\texttt{agent\_sit\_down}". The \texttt{navigate} action is non-atomic and can be further expanded into primitive operations such as ``\texttt{step\_forward}'' and ``\texttt{rotate}'' by executor. In the \textit{rule-based} execution mode, the planner computes the shortest path from the agent’s current position to the nearest chair, generating a sequence of navigation primitives such as \texttt{navigate(0, 1)}, \texttt{navigate(1, 10)}, and \texttt{navigate(10, 10)}, where \texttt{(10, 10)} denotes the chair’s location. Once the agent reaches the chair, the executor executes ``\texttt{agent\_sit\_down}'' and terminates when the action list becomes empty. In the \textit{visual-based} mode, the executor directly feeds environmental observations into a VLM (e.g., GPT-4o), which determines the next action step by step (e.g., execute ``\texttt{step\_forward}'' then ``\texttt{step\_forward}'' and finally ``\texttt{agent\_sit\_down}'') 
based on visual context.

\begin{figure}[H]
    \vspace{-10px}
    \centering
    \includegraphics[width=0.95\linewidth]{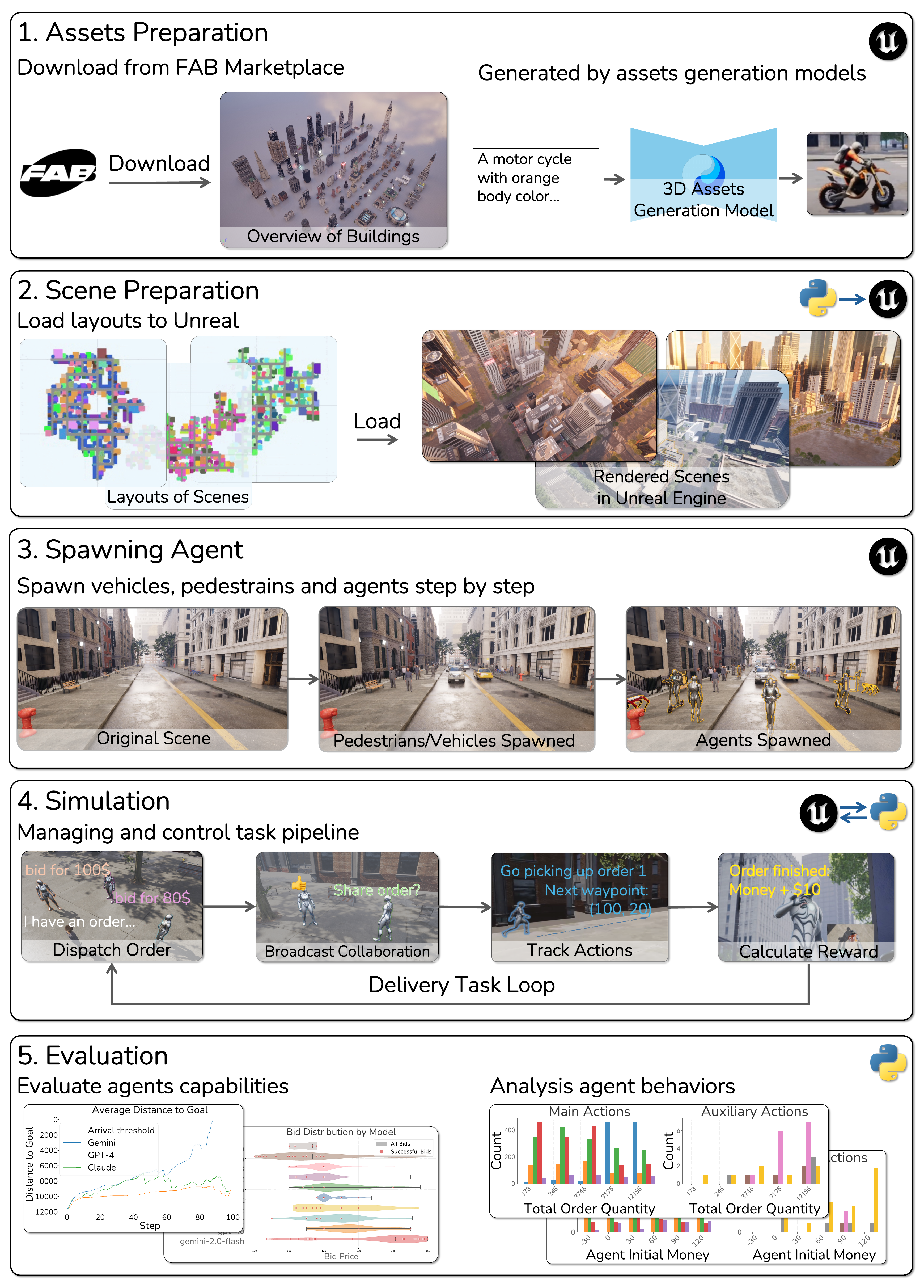}
    \vspace{-10px}
    \caption{\textbf{Workflow for Constructing a Task Suite in \simworld.}
    }\label{fig:workflow}
\end{figure}

\subsection{UnrealCV+ Communication Module}
\label{sec:unrealcv}
Inspired by UnrealCV~\cite{qiu2016unrealcv}, we develop UnrealCV+ as a communication module that bridges the Unreal Engine backend  (\S\ref{sec:ue_backend}) with the environment layer (\S\ref{sec:environment_layer}). By establishing a Transmission Control Protocol (TCP) connection, UnrealCV+ enables efficient, reliable, and bidirectional communication between the two sides. Implemented in both Python and C++, the module supports flexible data exchange and fine-grained control of the simulation.

To adapt UnrealCV+ to agent-task scenarios, we introduce a customized command set for scene control, actor manipulation, and data querying. For example, the environment layer can issue commands to the Unreal Engine backend such as ``\texttt{spawn actors at locations}'', ``\texttt{get position of a pedestrian}'', or ``\texttt{execute action of a robot}''.

Within the simulation loop, the environment layer governs the logical evolution of the simulation, while the Unreal Engine backend continuously returns updated physical states and visual observations of all agents. All communications are transmitted through UnrealCV+. This decoupled architecture cleanly separates logic computation from rendering, improving flexibility, scalability, and modularity.

\section{Case Study: Delivery Task}

\begin{figure}[t]
    \centering
    \includegraphics[width=\linewidth]{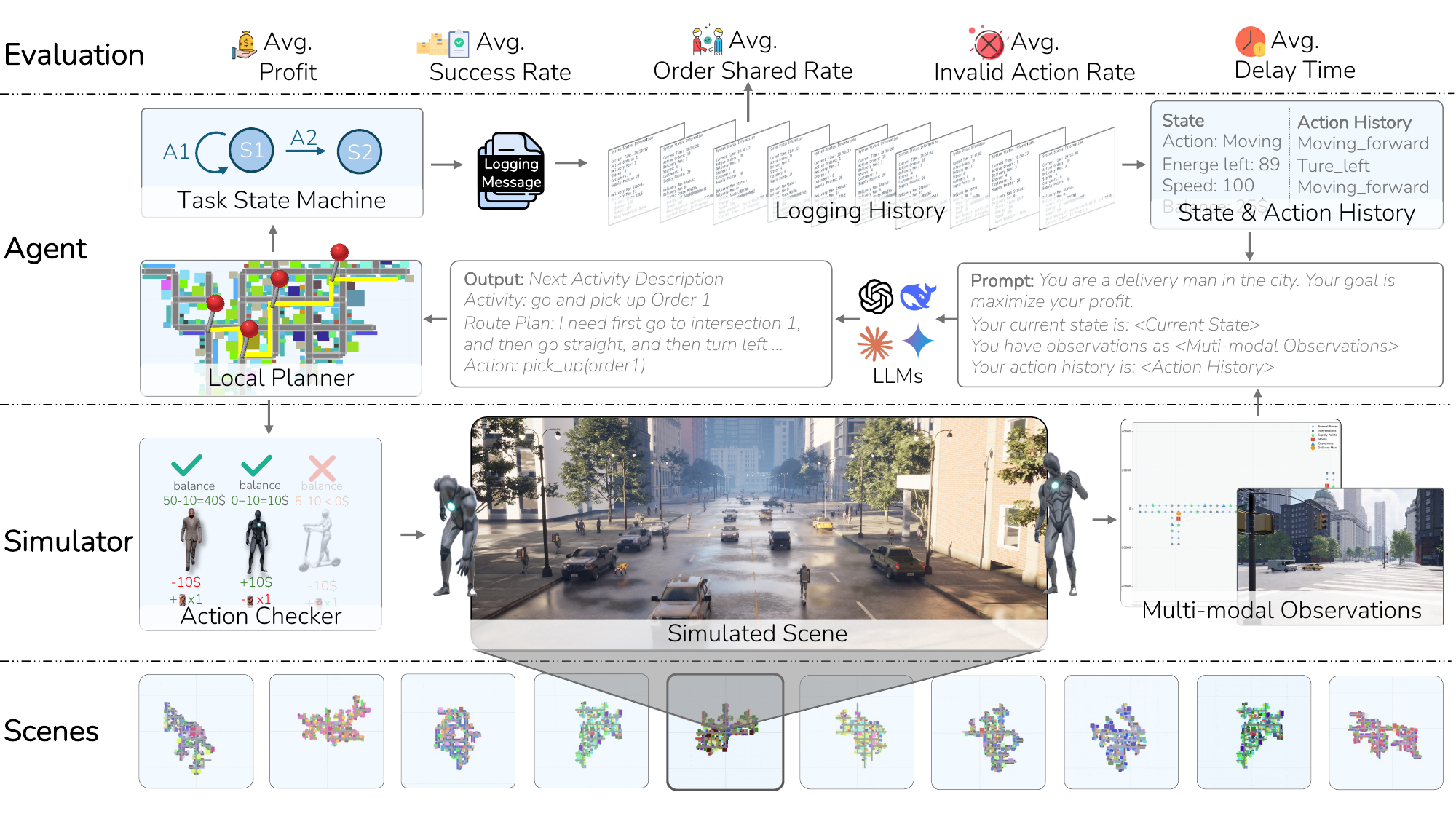}
    \caption{\textbf{Architecture of the Delivery Task.}}
    \label{fig:arc_casestudy}
\end{figure}

\simworld is built with extensibility and evaluation task creation as core design goals. It provides tools to easily define tasks, agent roles, reward functions, and evaluation metrics, minimizing the engineering effort required to create new experiments. Figure~\ref{fig:workflow} showcases how to define a multi-agent delivery task within the \simworld ecosystem. The process is delineated as follows:

First, users prepare assets for the delivery task, either by importing 3D assets purchased from third-party marketplaces or by generating them using text-to-3D models. Next, \simworld initializes the city environments via its procedural city generation module to build roads, buildings, trees, and city props. Third, vehicles, pedestrians and delivery agents are spawned. Fourth, the simulation runs, allowing delivery agents to act and interact within the configured environment dynamics. Finally, delivery agents are evaluated across multiple dimensions, such as delivery success rate, average completion time, and total profit. The following sections provide detailed definitions and analysis. The eventual architecture of the delivery task is shown in Figure \ref{fig:arc_casestudy}.


\begin{figure}[t]
    \centering
    \includegraphics[width=\linewidth]{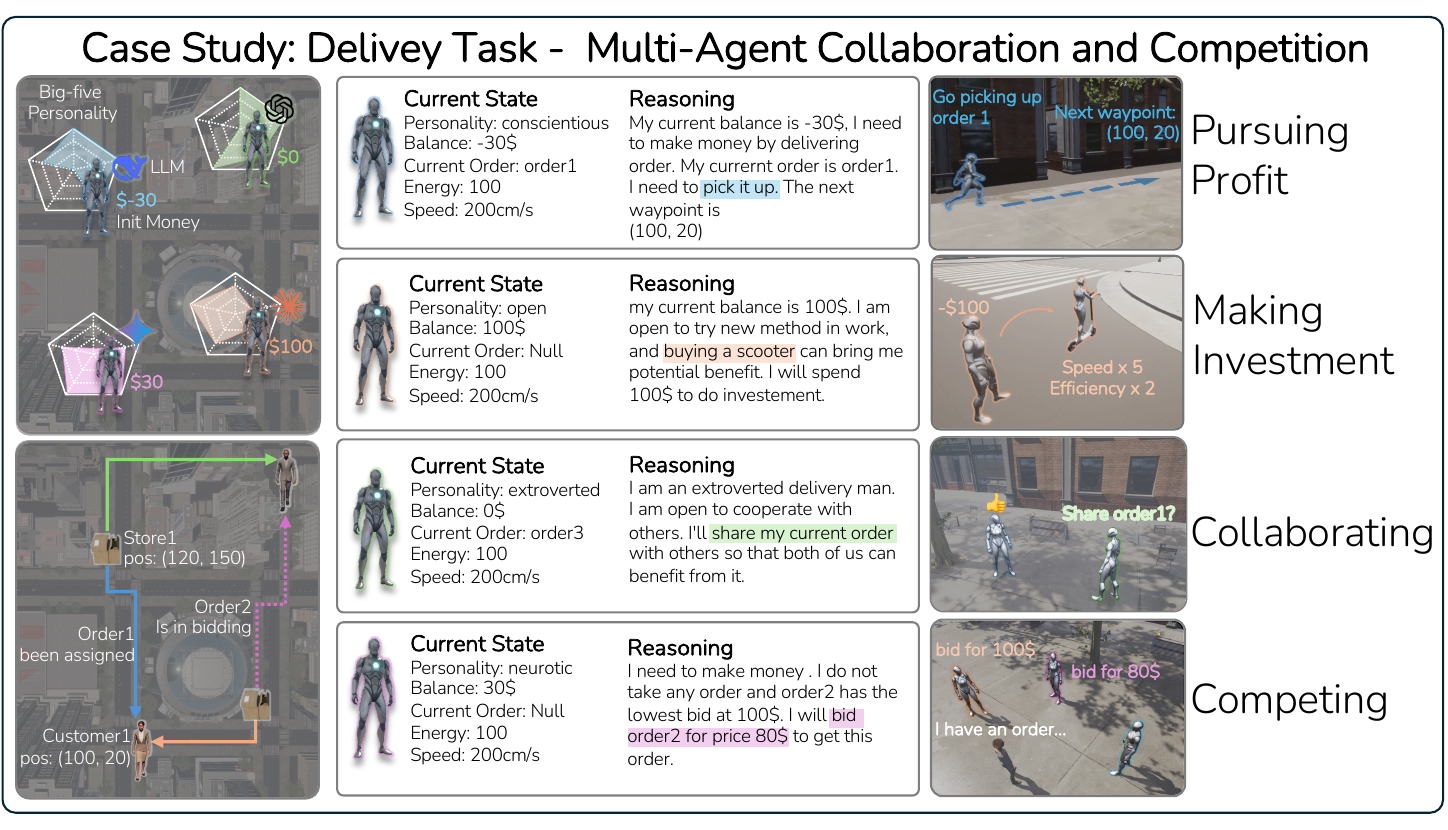}
    \caption{\textbf{Delivery Task.} A delivery scenario requiring multi-agent collaboration and competition. Each agent is initialized with distinct personalities and internal states and can act to grow, thrive, and ultimately maximize its earnings.}
    \label{fig:casestudy}
\end{figure}

\subsection{Task Formulation}
The delivery task is designed to evaluate the social reasoning capabilities of foundation models in realistic, open-world urban environments.
LLM agents are deployed as delivery agents in a city-scale environment built using \simworld. Their goal is to grow and thrive in this dynamic setting. To enhance realism and complexity, the environment incorporates several dynamic systems:
(1) an energy system, where agents must manage stamina and replenish it through consumables;
(2) an economic system, where agents earn and spend currency on purchases such as scooters and drinks;
(3) an order-sharing mechanism, enabling agents to collaborate by sharing delivery tasks and optimizing group performance.
These components create a rich, interactive simulation environment for evaluating agents’ decision-making, adaptability, and social reasoning in complex urban scenarios. The overview of the delivery task is shown in Figure \ref{fig:casestudy}.

\paragraph{Environment.}
All experiments are conducted on the same map generated by the procedural city generation module in \simworld. To ensure fair and efficient evaluation, graphical rendering is disabled during simulation. However, the physical simulation and evaluation modules remain active to preserve environment dynamics and task fidelity.

\paragraph{Action Space.}
The task features a two-tiered action space: high-level actions decided by LLM agents and low-level actions executed by the \simworld action planner. At each decision step, the agent may choose: \textbf{Bid Order} (offer a price to compete for a new order), \textbf{Pick Up Order} (navigate to the pick-up point), \textbf{Deliver Order} (complete delivery at the destination), \textbf{Share Order} (publish the order for collaboration), \textbf{Purchase Scooter} (purchase a scooter and equip it to move faster). The full action space is shown in Table \ref{tab:delivery-action-space}.

\begin{table}[t]
    \centering
    \caption{\textbf{Hierarchical Action Space Design in Delivery Task.} High-level actions are given to language models to make decision, which correspond to strategic decisions, while low-level actions are only exposed to local action planner module to execute concrete movements and interactions.}
    \label{tab:delivery-action-space}
    \resizebox{\textwidth}{!}{%
    \begin{tabular}{llll}
        \toprule
        \textbf{Action Level} & \textbf{Action Name} & \textbf{Description} &\textbf{Invocation Method} \\
        \midrule
        \multirow{9}{*}{High-Level} 
            & Bid Order & Offer a price to a new order on platform to compete with other Model Generations & \multirow{9}{*}{Model Generation} \\
            & Pick Up Order & Navigate to the pick-up point of order\\
            & Deliver Order & Navigate to the delivery point and complete the order\\
            & Share Order & Publish the order for multi-model generation cooperation\\
            & Cancel Share Order & Cancel a shared order that has been published\\
            & Go to Meet-point & Navigate to the meet point for the shared orde \\
            & Purchase Scooter & Buy and use a scooter \\
            & Purchase Drinks & Buy consumables to restore energy \\
            & Adjust Speed & Adjust travel speed\\
        \midrule
        \multirow{5}{*}{Low-Level}
            & Move Forward & Basic movement action & \multirow{5}{*}{Action Planner} \\
            & Stop & Stop moving \\
            & Rotate & Adjust the facing direction \\
            & Change Speed & Adjust walking speed \\
            & Drive Scooter & Control a scooter for movement  \\

        \bottomrule
    \end{tabular}
    }
\end{table}

\paragraph{Baseline Agents.}
We evaluate multiple foundation models serving as the backbone of delivery agents, including \texttt{Claude-3.5-Sonnet}, \texttt{DeepSeek-V3}, \texttt{GPT-4o}, \texttt{Gemini-2.5-Flash}, and \texttt{QWQ}. The ReAct~\cite{yao2022react} prompting framework is employed to explicitly separate reasoning and action selection.

\paragraph{Metrics.} 

Aligned with the agent’s hierarchical decision-making, we design a three-level evaluation framework. Overall performance is measured by \textbf{total profit} (the cumulative monetary gain the agent achieves over the simulation period), while operational effectiveness is assessed using \textbf{order success rate} (the proportion of orders that the agent successfully completes relative to the total assigned orders), \textbf{energy efficiency} (the ratio of energy consumed to revenue generated), \textbf{order sharing count} (the number of sharing orders), and \textbf{investment count} (the number of strategic investments).






\subsection{Main Results}
\label{sec4.2}
For each evaluation, we sample 20 agents controlled by the same language model, each running for 5000 simulation steps. At each step, an agent issues two API requests, averaging around 7000 tokens per request. Based on results in Table \ref{tab:main_results}, our empirical analysis over three simulation rounds reveals distinct operational behaviors across models.

\begin{table}[ht]
\centering
    \caption{\textbf{Performance of Model-Controlled Agents.} Metrics are reported as mean (Avg) and standard deviation (Std) over three 5000-step simulations. Bold indicates the best Avg per column.}
    \scalebox{1}{
    \begingroup
    \renewcommand{\arraystretch}{1}
    \setlength{\tabcolsep}{3pt}
    \hspace{-10pt}
    \begin{tabular}{lcccccccccc}
        \toprule
        \textbf{Model} 
        & \multicolumn{2}{c}{\cellcolor[HTML]{F2F6FE}\textbf{Profit}} 
        & \multicolumn{2}{c}{\cellcolor[HTML]{F2F6FE}\textbf{Successful Orders}} 
        & \multicolumn{2}{c}{\cellcolor[HTML]{F2F6FE}\textbf{Energy Efficiency}} 
        & \multicolumn{2}{c}{\cellcolor[HTML]{F2F6FE}\textbf{Sharing Count}} 
        & \multicolumn{2}{c}{\cellcolor[HTML]{F2F6FE}\textbf{Investment Count}} \\
        \cmidrule(lr){2-3} \cmidrule(lr){4-5} \cmidrule(lr){6-7} \cmidrule(lr){8-9} \cmidrule(lr){10-11}
        & Avg & Std & Avg & Std & Avg & Std & Avg & Std & Avg & Std \\
        \midrule
        DeepSeek-V3 & \textbf{69.48} & 16.77 & 2.10 & 0.47 & 0.34 & 0.07 & 2.33 & 0.47 & 8.00 & 3.00 \\
        Claude-3.5-Sonnet & 69.07 & 20.69 & \textbf{2.73} & 1.10 & \textbf{0.54} & 0.20 & \textbf{11.33} & 8.39 & \textbf{9.00} & 3.46 \\
        GPT-4o                     & 43.91 & 14.16 & 1.63 & 0.43 & 0.30 & 0.06 & 0.67 & 0.47 & 4.67 & 0.47 \\
        Gemini-2.5-Flash           & 42.42 & 3.10  & 2.10 & 0.17 & 0.17 & 0.04 & 2.67 & 1.25 & 2.00 & 2.00 \\
        Gemini-2.0-Flash           & 28.72 & 12.04 & 1.53 & 0.58 & 0.11 & 0.03 & 0.67 & 0.47 & 0.67 & 1.00 \\
        Qwen3-32B                  & 24.73 & 7.95  & 1.37 & 0.13 & 0.40 & 0.17 & 1.33 & 0.47 & 5.33 & 2.06 \\
        DeepSeek-Prover-V2         & 21.66 & 7.18  & 0.67 & 0.14 & 0.42 & 0.03 & 7.33 & 8.39 & 1.00 & 1.00 \\
        QwQ                        & 17.31 & 4.07  & 0.87 & 0.20 & 0.41 & 0.20 & 0.33 & 0.47 & 3.33 & 2.52 \\
        GPT-4o-mini                & 0.00  & 0.00  & 0.00 & 0.00 & 0.00 & 0.00 & 0.00 & 0.00 & 0.00 & 0.00 \\
        \bottomrule
    \end{tabular}
    \endgroup}
    \vspace*{-10pt}
    \label{tab:main_results}
\end{table}

\texttt{DeepSeek-V3} ($69.475 \pm 16.772$) and \texttt{Claude-3.5-Sonnet} ($69.068 \pm 20.685$) achieved the highest mean profits, with \texttt{Claude-3.5-Sonnet} also leading in mean successful orders ($2.733 \pm 1.102$) and energy efficiency ($0.5411 \pm 0.1981$). 
Notably, these superior average outcomes were associated with substantial performance variability, as reflected in their respective standard deviations. 

Conversely, \texttt{Gemini-2.5-Flash}, while attaining a moderate mean profit of $42.423$, exhibited markedly more consistent profit generation with a standard deviation of $\pm 3.103$, and also demonstrated stability in successful orders ($2.100 \pm 0.173$). 
Extreme variability was evident in specific metrics for certain models; for instance, sharing counts for \texttt{DeepSeek-Prover-V2} ($7.333 \pm 8.386$) and \texttt{Claude-3.5-Sonnet} ($11.333 \pm 8.386$) showed standard deviations exceeding their means, indicating highly unpredictable behavior in this aspect. 

The \texttt{GPT-4o-mini} model consistently yielded zero values across all metrics ($0.000 \pm 0.000$), suggesting it does not truly understand the goals well enough to make reasonable decisions based on the given instructions and context.

Higher investment strategies, such as those adopted by \texttt{DeepSeek-V3} ($8.000 \pm 3.000$) and \texttt{Claude-3.5-Sonnet} ($9.000 \pm 3.464$), generally correlated with greater mean profit achievement but also with increased outcome volatility. 
These findings underscore a prevalent trade-off between optimizing for peak average performance metrics and ensuring consistent, predictable agent behavior, a critical consideration for robust deployment in dynamic environments.

\begin{AIbox}{Takeaway: Model Performances in Multi-agent Tasks}
Top-performing models such as \texttt{DeepSeek-V3} and \texttt{Claude-3.5-Sonnet} achieve high average profits but show greater variance, whereas \texttt{Gemini-2.5-Flash} demonstrates more consistent yet moderate performance. \texttt{GPT-4o-mini} failed entirely across all metrics. Overall, the results highlight a trade-off between maximizing average performance and ensuring consistent, reliable behavior (Table~\ref{tab:main_results}).
\end{AIbox}

\subsection{Ablation Study}
\label{sec4.3}
To take a step deeper on multi-agent collaboration and competition, we conduct three ablation experiments. In the \textbf{Model Competition} setting, we sample 24 agents controlled by 12 models, with each model managing two agents over 1000 rounds. In this ablation experiment, we study how models make choices within a highly competitive environment in order to maximize their returns; 
In the \textbf{Environment Configuration} setting, we vary two environment configurations: the initial financial budget and the global order volume. For each configuration, we sample several stages from low to high to observe how agents’ behavior changes with the environment conditions;
In the \textbf{Persona} setting, we use the model with the best performance to control the agents with persona description in prompts. We sample 20 agents, assign each persona to two agents, and observe how these personas shape the agents’ behaviors and decision strategies.

\paragraph{Model Competition.}
To intensify inter-agent competition, we constrain each agent to handle at most one order at a time and set the environment's hunger rate to 0.9, ensuring a high demand for delivery. In each experimental session, 24 agents are jointly controlled by 12 different models, where each model governs two agents. These agents actively bid for orders in a shared environment with the goal of maximizing profit. Each session runs for 1000 simulation steps, and results are averaged across three random seeds.

\begin{figure}[ht]
    \centering
    \begin{minipage}{0.55\linewidth}
        \centering
        \includegraphics[width=\linewidth]{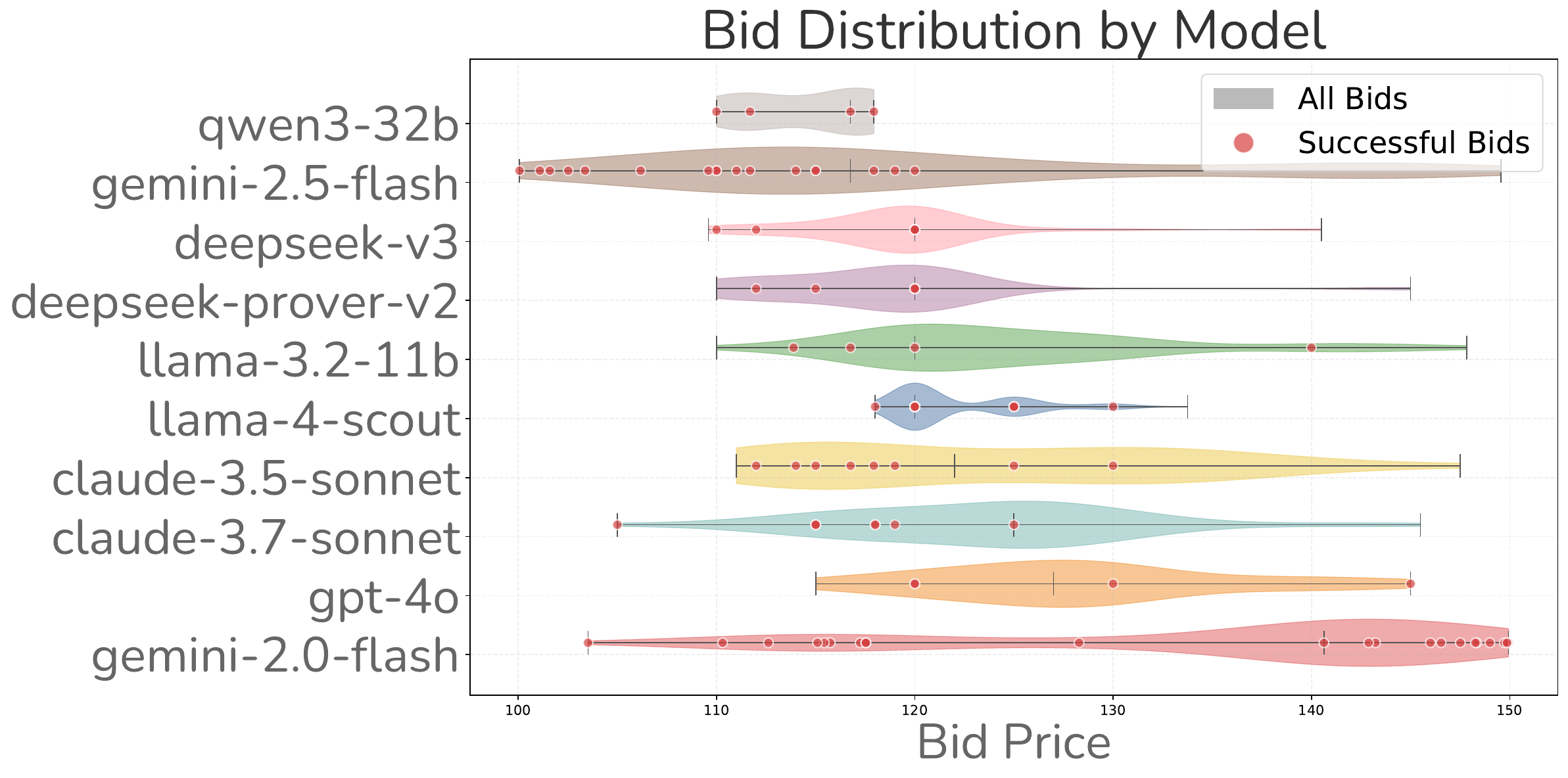}
        \subcaption{Bid price distribution by model. The violin plots illustrate the distribution of bid prices generated by each model. Red points indicate bids that were successfully accepted.}
        \label{fig:bid(a)}
    \end{minipage}
    \hfill
    \begin{minipage}{0.4\linewidth}
        \centering
        \includegraphics[width=\linewidth]{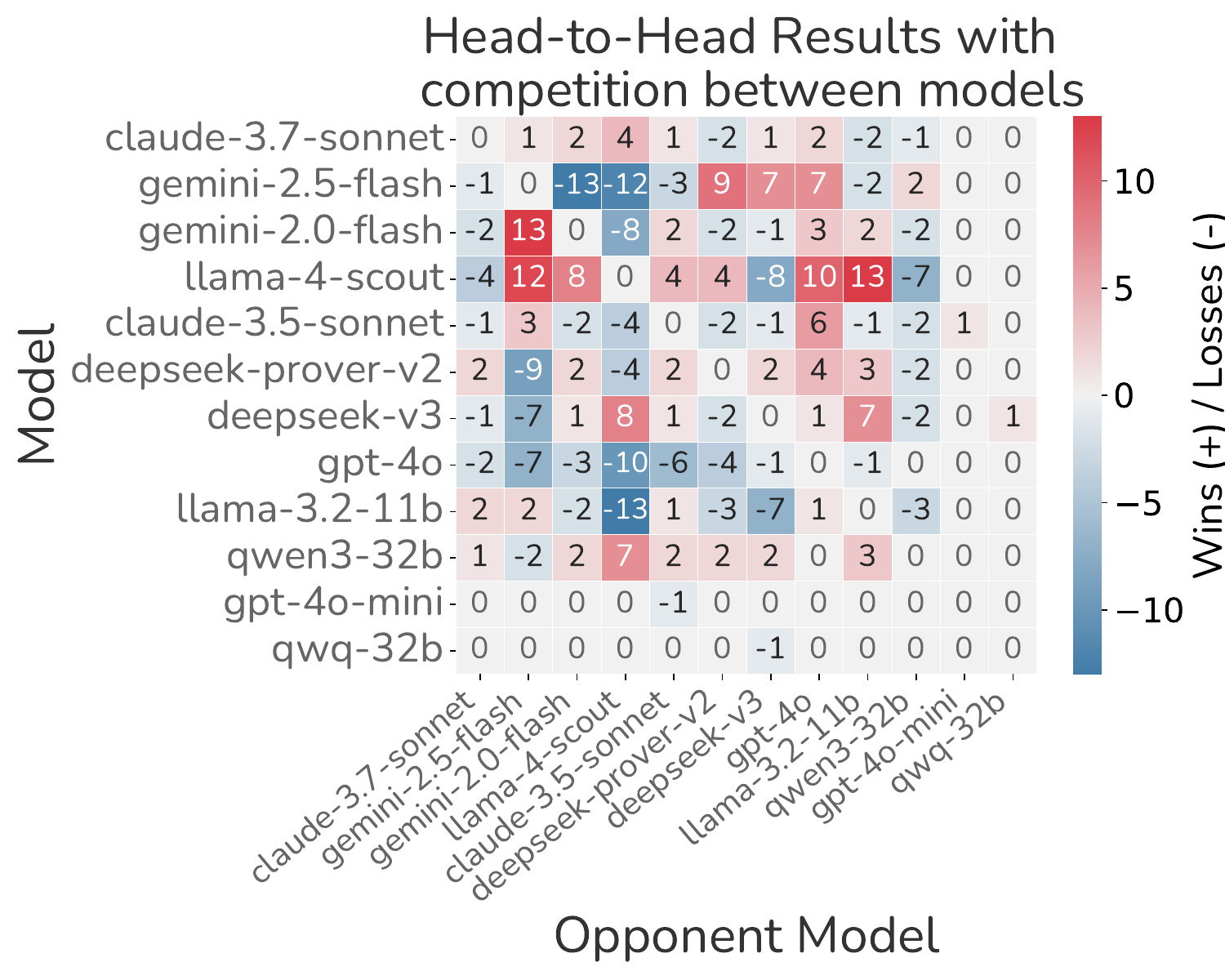}
        \subcaption{Win-Loss Matrix of Model Competition.}
        \label{fig:bid(b)} 
    \end{minipage}
    \captionsetup{justification=raggedright} 
    \vspace{-5pt}
    \caption{
        \textbf{Bidding Behavior and Evaluation Results.} 
        \textbf{(a)} 
        Lower bid prices may increase the likelihood of being assigned an order, but often come at the cost of reduced profit margins.
        \textbf{(b)} 
        Higher values in red represent more wins; lower values in blue indicate more frequent losses in pairs.\vspace{-2pt}
    }\label{fig:bid}
\end{figure}

As shown in Figure~\ref{fig:bid(a)}, models exhibit distinct bidding behaviors. Notably, \texttt{Claude-3.7-Sonnet}, \texttt{Gemini-2.5-Flash} and \texttt{Gemini-2.0-Flash} demonstrate broad bid price distributions, indicating a flexible bidding strategy. This flexibility increases their chances of winning orders when in competition with other models. In contrast, models such as \texttt{LLaMA-4-Scout} and \texttt{LLaMA-3.2-11b} tend to use narrower bidding ranges, which limits their competitiveness and results in lower win rates.

Figure~\ref{fig:bid(b)} presents the head-to-head competition outcomes. \texttt{Deepseek-Prover-V2} and \texttt{Qwen3-32B} achieve the highest win rates against other models. This is primarily because they often bid lower prices, making their offers more likely to be accepted by the platform. Conversely, models like \texttt{GPT-4o} and \texttt{LLaMA-3.2-11b} tend to place higher bids, reducing their success rate despite frequent participation. Models such as \texttt{QwQ-32B} and \texttt{GPT-4o-mini} are less active overall, leading to fewer bids and lower order acquisition rates. This inactivity contributes to their diminished final profit, as shown in Table~\ref{tab:main_results}.

\begin{AIbox}{Takeaway: Multi-Agent Competition}
Models with flexible bidding strategies, like \texttt{Claude-3.7-Sonnet} and \texttt{Gemini-2.5-Flash}, achieve higher order win rates, while those with narrow or high bids, like \texttt{LLaMA-3.2-11b} and \texttt{GPT-4o}, underperform. Models that bid aggressively, such as \texttt{Deepseek-Prover-V2} and \texttt{Qwen3-32B}, dominate head-to-head competitions, whereas inactive models like \texttt{GPT-4o-mini} fail to secure bids and profits.
\texttt{QwQ} and \texttt{GPT-4o-mini} show minimal bidding activity and weak task participation (Figure~\ref{fig:bid}).
\end{AIbox}

\paragraph{Environment Configuration.}
We further investigate how different environmental configurations impact agent behavior and overall performance. Specifically, we explore two key factors: the global order availability and the agents’ initial financial endowment. For each factor, we conduct a series of controlled experiments to observe how variations affect agents’ action distributions.

\begin{figure}[ht]
    \centering
    \includegraphics[width=\linewidth]{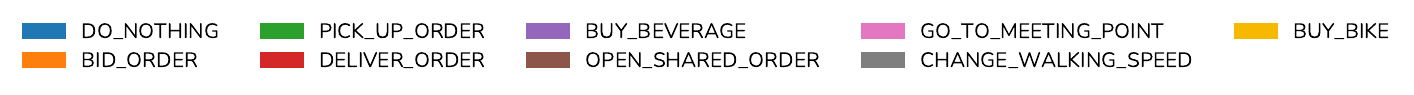}
    \begin{minipage}{0.49\linewidth}
        \centering
        \includegraphics[width=\linewidth]{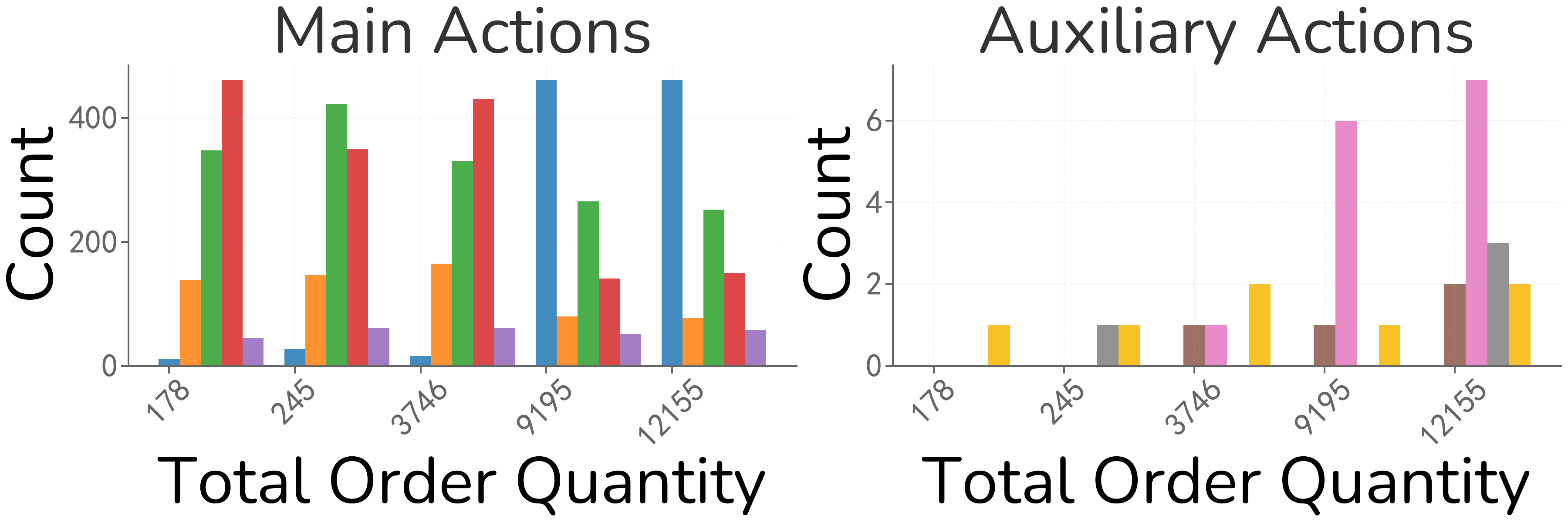}
        \subcaption{Effect of Order Quantity}
        \label{fig:environment(a)}
    \end{minipage}
    \hfill
    \begin{minipage}{0.49\linewidth}
        \centering
        \includegraphics[width=\linewidth]{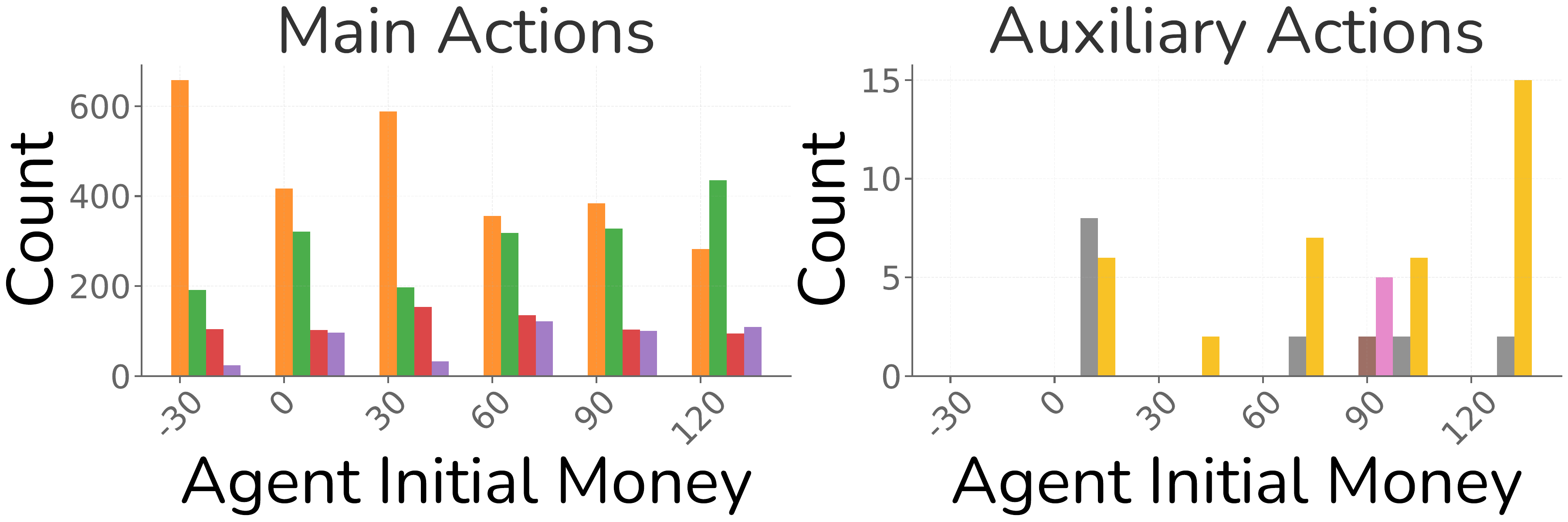}
        \subcaption{Effect of Initial Money}
        \label{fig:environment(b)}
    \end{minipage}
    \caption{\textbf{Action Distribution across Environmental Settings.} 
    (a) shows how global order quantity affects agent behavior; 
    (b) shows the effect of initial money on action selection.
    \label{fig:environment}}
\end{figure}

As shown in Figure~\ref{fig:environment(a)}, when the total number of available orders increases, agents tend to perform fewer pickup and delivery actions and instead choose the \texttt{do nothing} action more frequently. 
This suggests that in resource-rich environments, agents are more inclined to conserve energy and avoid unnecessary effort, opting to wait for optimal opportunities rather than actively pursue deliveries. 
Conversely, in low-resource settings, agents are more motivated to engage in delivery tasks to secure profits. 
Additionally, as resource abundance increases, agents demonstrate a higher tendency to initiate and complete shared deliveries, likely as a means to reduce energy costs through collaboration.

Figure~\ref{fig:environment(b)} illustrates the impact of agents’ initial monetary resources. As initial capital increases, agents are less reliant on aggressive bidding and instead prioritize actions such as order pickup. When funds are limited, competition intensifies, leading to more frequent bidding behavior. Furthermore, with sufficient initial capital, agents are more willing to invest in infrastructure, such as purchasing a scooter, which enhances their long-term delivery efficiency.

\begin{AIbox}{Takeaway: Resource and Decision-Making Strategy}
Order resource scarcity increases agent competitiveness and task urgency. Sufficient agent initial money leads to more relaxed, profit-insensitive behavior (Figure~\ref{fig:environment}).

\end{AIbox}
These observations suggest that agents are more competitive and task-driven in resource-constrained environments. In contrast, resource-rich conditions reduce the urgency to complete tasks and generate immediate profits. Importantly, agents are also more likely to engage in actions that involve upfront costs but promise long-term benefits—such as investment and shared delivery—provided they have the financial capacity and enough orders taken to do so.

\begin{figure}[ht]
    \centering
    \begin{minipage}{0.42\linewidth}
        \centering
        \includegraphics[width=\linewidth]{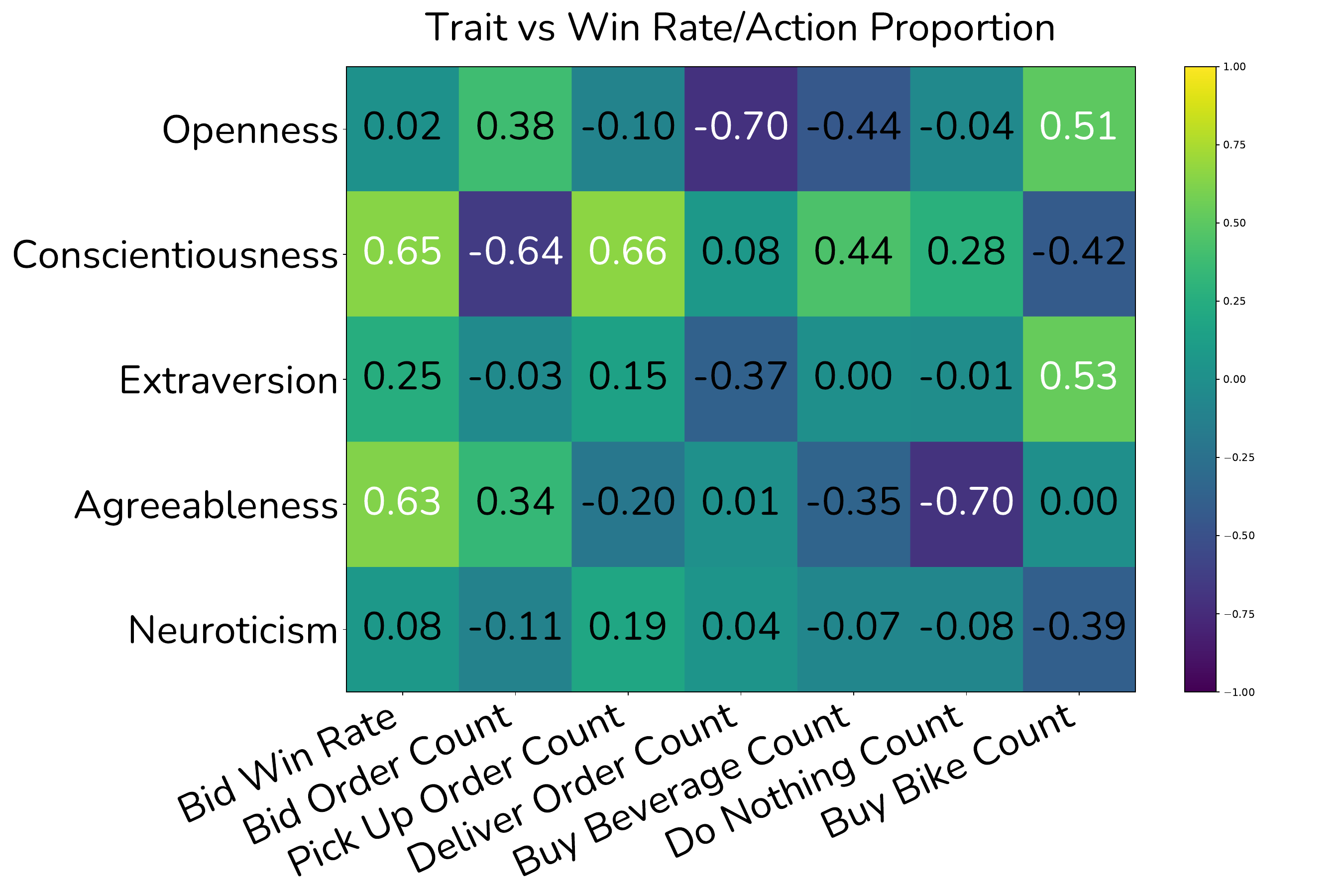}
        \subcaption{Pearson correlation b/w Big Five personality traits and agent behaviors.}
        \label{fig:persona(a)}
    \end{minipage}
    \hfill
    \begin{minipage}{0.55\linewidth}
        \centering
        \includegraphics[width=\linewidth]{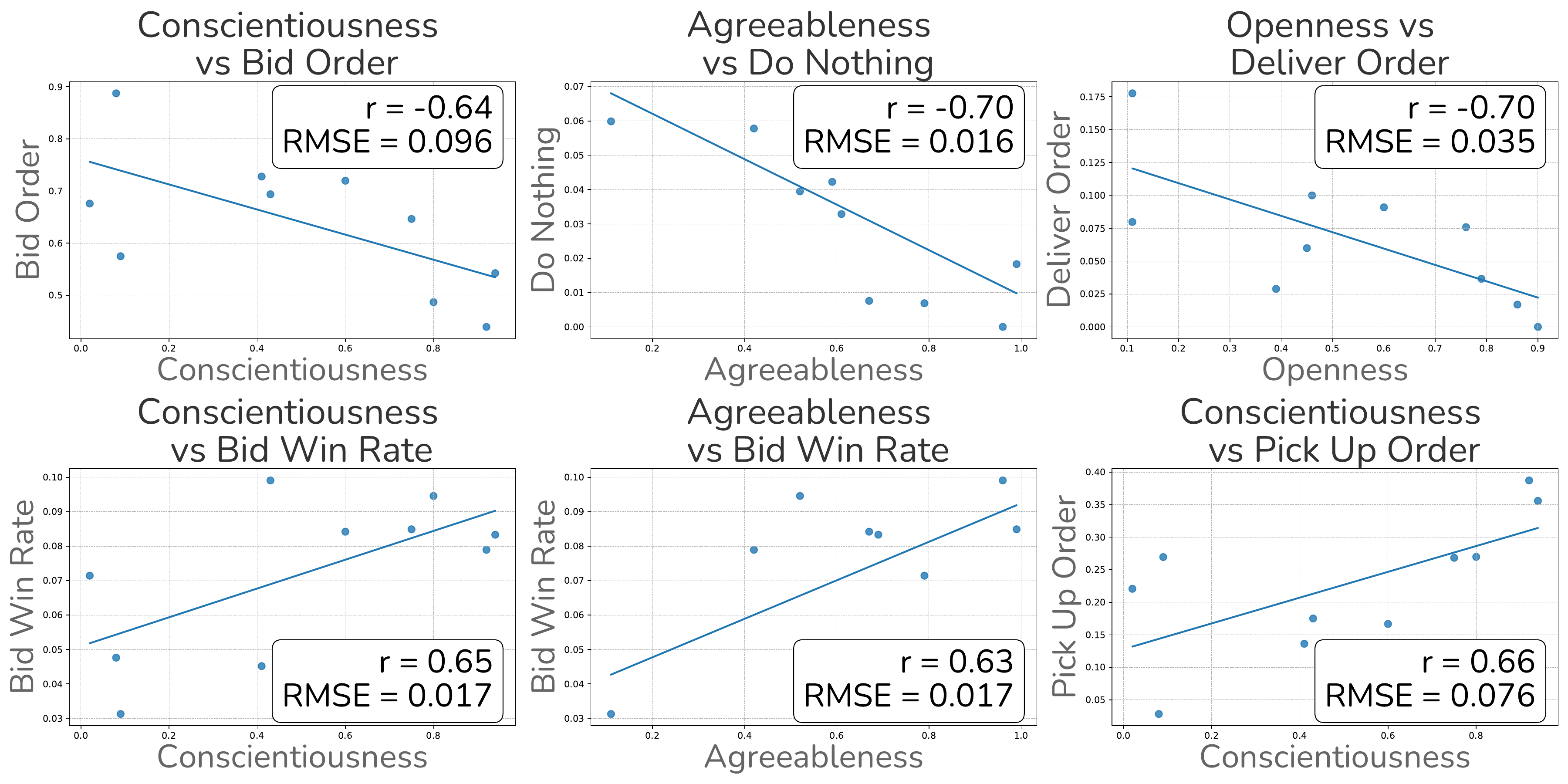}
        \subcaption{Linear Regression Reveals Strong Correlations between Behaviors and Persona Traits.}
        \label{fig:persona(b)} 
    \end{minipage}
    \captionsetup{justification=raggedright} 
    \vspace{-5pt}
    \caption{
        \textbf{Persona Influence on Agent Performance and Behavior.} 
        \textbf{(a)}
        Agents with higher Agreeableness are less likely to remain inactive and tend to achieve higher bid win rates. Conversely, agents with lower agreeable--
        ness display higher inactivity and narrower bidding price ranges, limiting their competitiveness. \textbf{(b)} The results demonstrate that agent behaviors are tightly aligned with their corresponding persona attributes, highlighting the effectiveness of trait-driven behavior modeling.
    }\label{fig:persona}
\end{figure}

\paragraph{Influence of Persona.}
Personality traits significantly affect the decision-making and performance of delivery agents. As shown in Figure~\ref{fig:persona}, agents with higher Conscientiousness tend to exhibit a lower frequency of bidding actions, a higher frequency of task-completion actions (e.g., picking up orders), and achieve a higher bid win rate. This suggests that conscientious agents prioritize task completion over strategic competition.
Agents with higher Agreeableness are less likely to remain inactive (i.e., perform \texttt{do nothing} actions) and tend to achieve higher bid win rates. Conversely, agents with lower Agreeableness display higher inactivity and narrower bidding price ranges, limiting their competitiveness.
Interestingly, agents with higher Openness exhibit reduced engagement in delivery tasks, possibly because they explore competitive or unconventional bidding strategies that divert attention from task execution.

\begin{AIbox}{Takeaway: Impact of Persona in Multi-agent Interaction}
Agent personalities shape strategic tendencies: conscientious agents prioritize task fulfillment, while openness and agreeableness modulate competitiveness and inactivity (Figure~\ref{fig:persona}).
\end{AIbox}

\section{Related Works}
Simulations have played a crucial role in constructing environments for training and evaluating autonomous agents. Text-based simulators often emphasize social scenarios, such as human interaction~\cite{yang2024oasis}, daily activities~\cite{park2023generative}, and relational polarization~\cite{piao2025agentsociety}. Popular embodied simulators support a broader range of applications, particularly in embodied AI research and 2D/3D scene synthesis~\cite{li2025simworld}. However, most embodied simulators remain constrained to either indoor household environments (e.g., AI2-THOR~\cite{kolve2017ai2}, Habitat~\cite{puig2023habitat}, iGibson~\cite{li2021igibson}) or outdoor driving scenarios (e.g., CARLA~\cite{2017carla}, MetaDrive~\cite{li2022metadrive}) or natural scenes (e.g., AirSim~\cite{shah2017airsim}). Most of these simulators~\cite{2017carla, puig2023habitat, li2021igibson, shah2017airsim, wang2024grutopia, gao2024embodiedcity} rely on a limited number of manually crafted scenes, which hinders scalability and diversity. Some platforms, such as MetaUrban~\cite{wu2025metaurban}, MetaDrive~\cite{li2022metadrive}, AI2-THOR~\cite{kolve2017ai2} and Genesis~\cite{Genesis}, introduce rule-based procedural generation to alleviate this issue. Nonetheless, existing embodied simulators typically lack support for dynamic multi-agent interactions in complex diverse environments.

Recent advancements have introduced large-scale, language-driven social simulators capable of modeling complex societal behaviors. OASIS~\cite{yang2024oasis} simulates up to one million LLM-powered agents interacting on social media platforms, capturing phenomena such as information diffusion, echo chambers, and polarization. Casevo~\cite{jiang2024casevo} integrates chain-of-thought reasoning, retrieval-augmented generation, and customizable memory mechanisms to simulate intricate social phenomena and decision-making processes. MineLand~\cite{yu2024mineland} offers a multi-agent Minecraft environment where agents, driven by physiological needs and limited multimodal perception, engage in collective behaviors, fostering ecological and detailed simulations. Project sid~\cite{al2024projectsid} further advances this landscape by deploying a large number of AI agents within a Minecraft environment to explore the emergence of AI civilizations. VirtualCommunity~\cite{zhou2025virtual} leverages Genesis simulator to conduct community influence task in outdoor, multi-agent scenes. These simulation platforms demonstrate agents' capabilities to form complex social structures, economies, and governance systems, providing insights into large-scale societal simulations and agentic organizational intelligence.

None of the existing simulators are explicitly designed to support dynamic, multi-agent interactions in large-scale outdoor and other diverse environments with both realistic rendering and physical simulation. \simworld addresses this limitation by providing a scalable, procedurally generated, and LLM/VLM-compatible platform that enables multi-agent collaboration and competition, language-grounded interactions, and comprehensive benchmarking for embodied intelligence.

Another emerging direction in world simulation involves end-to-end neural world models, which generate interactive video predictions conditioned on environment states, agent actions, and high-level controls \citep{xiang2024pandora, genie32025, xiang2025pan}. Recent systems can simulate short video rollouts or 3D-consistent scenes using learned dynamics, offering a flexible alternative to traditional engine-based simulation. 
On the other hand, \simworld, built on the Unreal Engine, provides high-fidelity, physically grounded, and deterministically controllable environments capable of supporting thousands to even millions of interacting agents at scale. Moreover, because \simworld supports diverse, high-quality procedural and handcrafted scenes, it can serve as a powerful generator of large-scale training data, offering a rich source of supervised trajectories, multi-agent interactions, and physically realistic rollouts that can be used to train and improve neural world models.

\bibliography{references_agent,references_robot}
\bibliographystyle{icml2025}


\end{document}